\documentclass[runningheads]{llncs}

\usepackage{eccv}
\usepackage{eccvabbrv}
\usepackage{graphicx}
\usepackage{booktabs}
\usepackage[accsupp]{axessibility}

\usepackage[pagebackref,breaklinks,colorlinks,citecolor=eccvblue]{hyperref}
\usepackage{hyperref}
\usepackage{algorithm}
\usepackage{algpseudocode}

\usepackage{capt-of}
\usepackage{orcidlink}
\usepackage{pifont}
\usepackage{float} 
\usepackage{comment}
\usepackage{multirow}   
\usepackage{xcolor}     
\usepackage[table]{xcolor}
\newcommand{\Letter}{\ding{41}}

\begin{document}

\title{FoundationGeo: Learning Spatial Pixel-Wise Fields for Monocular Metric Geometry} 

\titlerunning{FoundationGeo}




\author{
Muxin Liu\inst{1,2}$^{*}$ \and
Xiaoyang Lyu\inst{1}$^{*}$ \and
Tianhe Ren\inst{1} \and
Peng Dai\inst{1} \and \\[0.2em]
Xiaoshan Wu\inst{1}  \and
Zhiyue Zhang\inst{1} \and
Jiaqi Zhang\inst{1} \and
Jiehong Lin\inst{1} \and \\[0.2em]
Shaoshuai Shi\inst{2}\textsuperscript{\Letter} \and
Xiaojuan Qi\inst{1}\textsuperscript{\Letter}
}

\authorrunning{M.~Liu et al.}

\institute{
The University of Hong Kong
\and
Voyager Research, DiDi Chuxing \\[2pt]
\small
$^{*}$ Equal contribution
\quad
\textsuperscript{\Letter} Corresponding author
\quad
\email{\{mxliu,shawlyu\}@connect.hku.hk, 
shaoshuaics@gmail.com, 
xjqi@eee.hku.hk}
}

\maketitle

\begin{center}
    \vspace{-14pt}
    \centering
    \includegraphics[width=0.99\textwidth]{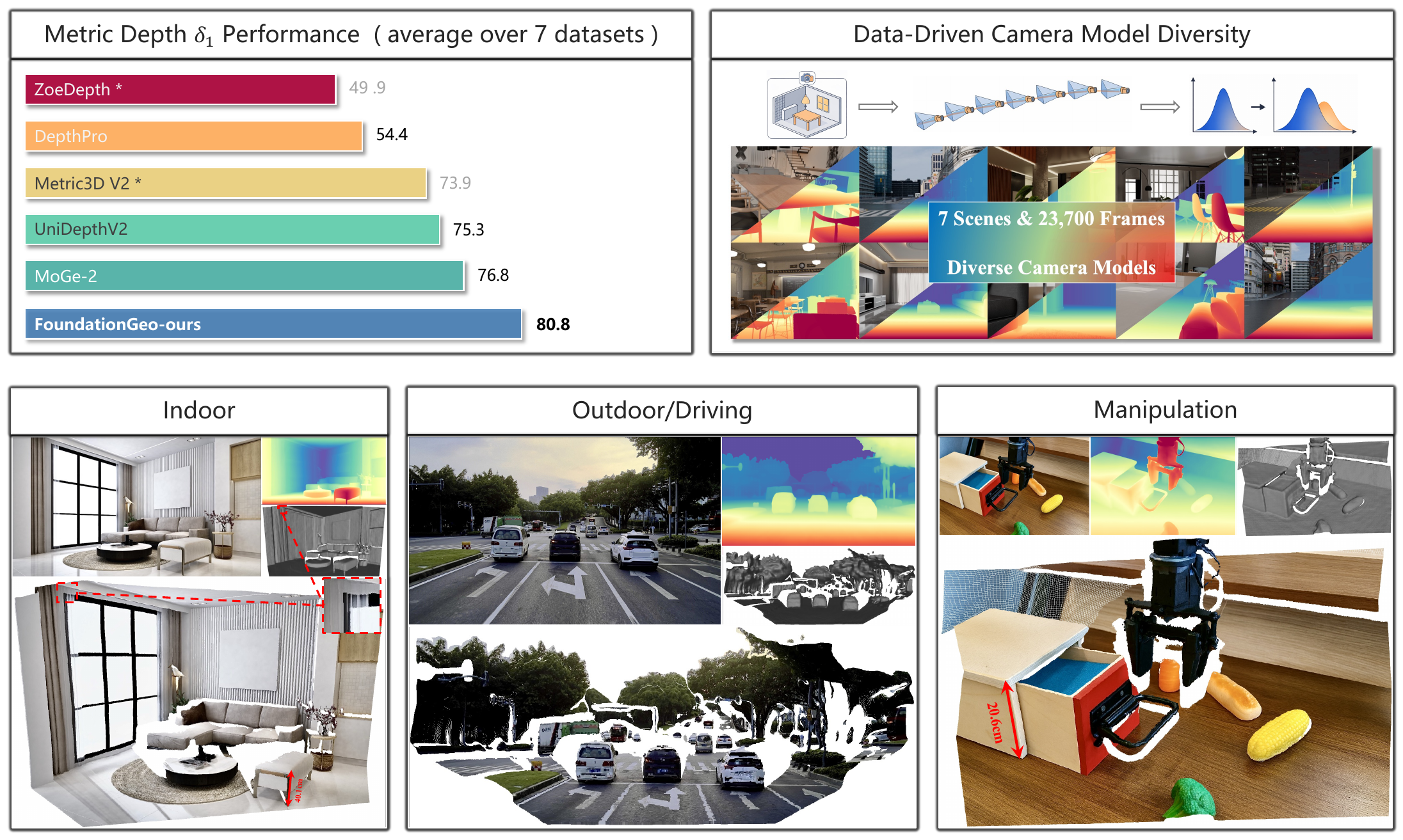}
    \captionof{figure}{Given an input image, our method recovers the metric 3D geometry of the scene, producing high-quality reconstructions that generalize well to open-domain data. }
    \label{fig:vis}
    \vspace{-14pt}
\end{center}%

\begin{abstract}
We present FoundationGeo, a two-stage framework that explicitly bridges relative and metric prediction via spatial calibration and principled data design. Stage 1 learns a high-fidelity, affine-invariant geometry model by initializing with DINOv3 and training on a curated 10.2M-sample multi-domain corpus with complementary local–detail supervision, yielding sharp boundaries and strong cross-domain generalization. Stage 2 moves beyond global scaling by introducing lightweight pixel-wise calibration fields for metric estimation: a scale field for spatially varying metric alignment and a ray-direction correction field that mitigates directional bias in point-map geometry, together producing metrically consistent 3D point maps.
Beyond model design, we identify camera intrinsic coverage, especially focal length distribution mismatch between training and test data, as a key bottleneck for zero-shot metric generalization: performance drops sharply when test intrinsics fall outside the training distribution. To address this, we synthesize additional training data across diverse focal lengths using a Blender-based data engine, repairing under-covered focal regimes and improving robustness under intrinsic shift.
Extensive zero-shot evaluations across seven benchmarks show that FoundationGeo significantly strengthens cross-domain robustness, staying near the top across diverse domains while avoiding the sharp cross-domain performance drops observed in other methods. This consistency translates into the best overall performance, surpassing heavier baselines by over 5.2\% on average. \noindent\textbf{Project page:} \url{https://mx-liu6.github.io/FoundationGeo-web/}

\label{abstract}

\end{abstract}
    
\section{Introduction}
\label{sec:intro}

Depth estimation from a single RGB image~\cite{saxena2008make3d,eigen2014depth} is a long-standing problem in computer vision, underpinning diverse applications such as 3D reconstruction, robotics, and AR/VR. Depending on the output representation, existing methods can be broadly categorized into relative and metric depth estimation. Relative approaches~\cite{yang2024depth,yang2024depth2,wang2025moge,ke2025marigold} predict geometry up to an affine ambiguity, effectively recovering accurate local shape and ordering without physical scale. In contrast, metric depth estimation~\cite{bhat2023zoedepth,piccinelli2024unidepth,piccinelli2025unidepthv2,hu2024metric3d,yin2023metric3d,wang2025moge2,bochkovskiydepth} aims to infer scale-aware geometry consistent with real-world units, which is critical for tasks requiring measurement, interaction, or physical reasoning. However, recovering metric scale from monocular input is intrinsically ill-posed due to perspective projection ambiguities, the dependence on camera intrinsics and scene priors. As a result, relative depth models have achieved much higher accuracy and generalization, while monocular metric depth estimation still lags far behind.

Existing metric solutions broadly fall into two families.
(1) \textit{Camera-model–based methods}~\cite{yin2023metric3d,hu2024metric3d,bochkovskiydepth,piccinelli2024unidepth,piccinelli2025unidepthv2} recover metric scale by explicitly modeling camera intrinsics, either by normalizing them into a canonical space~\cite{yin2023metric3d,hu2024metric3d} or by predicting focal length~\cite{bochkovskiydepth}. While this improves camera awareness, performance can be fragile under intrinsic miscalibration and sensitive to domain shifts across camera models.
(2) \textit{Relative-to-metric methods}~\cite{bhat2023zoedepth,wang2025moge2} repurpose strong relative predictors via a lightweight metric calibration module. A representative example is MoGe-2~\cite{wang2025moge2}, which converts high-quality relative predictions into metric estimates using a single global scale, thereby inheriting strong priors from relative branch and preserving fine details. This direction is appealing because it leverages robust relative geometry while requiring far less metric supervision than training a metric model from scratch.

\begin{figure}[ht]
    \centering
    \vspace{-12pt}
    \includegraphics[width=0.99\linewidth]{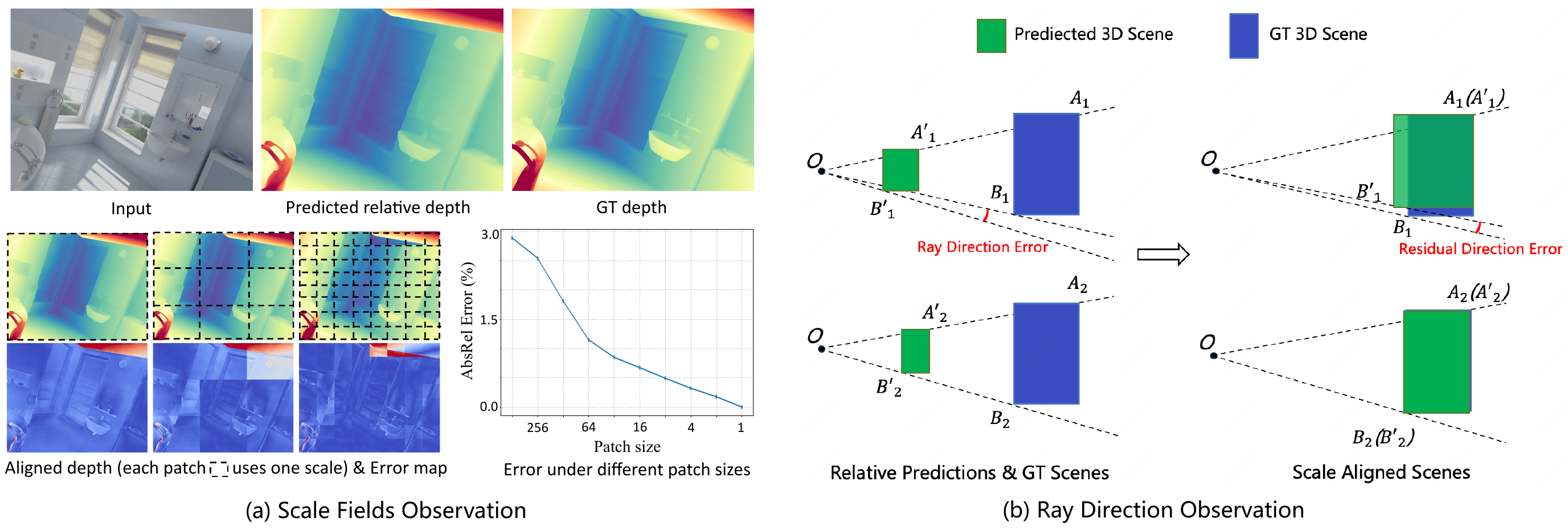}
    \caption{Observations on the relative to metric gap under point map supervision. (a) Scale misalignment is strongly spatially varying: as local scale alignment becomes increasingly patchified from coarse regions to finer patches, errors are corrected more effectively, and the global AbsRel (\%) decreases monotonically toward the per pixel limit, indicating the need for pixel wise calibration rather than a single global scale. (b) Beyond scale, metric errors also contain ray direction bias: even after scale is well aligned, residual directional error leads to unavoidable 3D inconsistency in the predicted point map, motivating an explicit ray direction correction field in addition to a pixel wise scale field.}
    \vspace{-16pt}
    \label{fig:Observation}
\end{figure}

Motivated by this landscape, we revisit relative-to-metric transfer and ask how far it can be pushed to close the remaining gap. Specifically, we investigate three questions.
\textbf{(a) How far can relative pretraining go in benefiting metric depth?} If the relative backbone becomes sufficiently strong and diverse, does metric calibration become the dominant bottleneck, or do other factors limit performance?
\textbf{(b) What is the right way to learn scale for metric prediction?} MoGe-2 shows that a single global scale can already yield strong metric outputs, but our observations reveal two fundamental limitations. First, Fig.~\ref{fig:Observation}(a) shows that as scale alignment becomes increasingly local (from coarse patches to finer ones), AbsRel decreases monotonically toward the per-pixel limit, indicating the need for \emph{spatially varying} calibration. Second, Fig.~\ref{fig:Observation}(b) exposes an orthogonal error source under point-map supervision: even with well-aligned scale, \emph{ray-direction bias} still induces 3D inconsistency.
\textbf{(c) What is the remaining gap on the data side, and how should we collect metric supervision effectively?} Beyond model design, we hypothesize that data remains a major bottleneck; closing the gap requires a clearer empirical understanding of what distributions matter for metric learning and principled guidance on how to collect or synthesize metric supervision to cover them effectively. 

Based on these findings, we propose \textbf{FoundationGeo} to close the relative-metric gap by jointly strengthening the relative foundation model and redesigning metric calibration under principled data-collection guidelines. First, we upgrade the relative backbone via DINOv3 initialization~\cite{simeoni2025dinov3}, large-scale training on a curated 10.2M-sample multi-domain corpus, and complementary \emph{local–detail} supervision with multi-scale feature fusion, yielding strong fine-detail fidelity and robust cross-domain generalization. Second, we move beyond global scaling with an explicit metric calibration module: we learn a pixel-wise scale field with an improved scale-map formulation and a ray-direction correction branch, together with tailored losses that disentangle and stabilize the learning of these metric components. In this stage, we optimize a joint objective that \emph{retains the relative-geometry loss as a structural regularizer} while learning metric calibration from limited but informative metric supervision. Third, we systematically analyze how training data coverage affects metric robustness (Fig.~\ref{fig:focal_study}). By quantifying errors across camera-intrinsic regimes, we show that performance drops sharply when test intrinsics fall outside the training distribution, identifying intrinsic diversity as a key driver of metric generalization. This motivates using intrinsic cues-- such as focal length and pose-- via intrinsic-conditioned sampling as a principled axis for data design and augmentation, explicitly targeting under-covered camera regimes rather than relying on indiscriminate dataset mixing. Concretely, we build a \emph{Blender-based data engine} to render synthetic images across diverse focal-length settings, repairing camera coverage and improving robustness.

Extensive zero-shot evaluations across seven datasets show that FoundationGeo substantially improves cross-domain robustness, remaining consistently competitive across diverse domains and stable under domain and camera-model shifts. This stability translates into the strongest overall metric depth performance. Compared with a prior method, FoundationGeo improves the averaged results from $15.7$ to $14.8$ in \textit{AbsRel} ($5.7\%$) and from $76.8$ to $80.8$ in $\delta_1$ ($5.2\%$). In addition, our first-stage FoundationGeo base model also achieves strong performance on relative depth accuracy and boundary F1, indicating that it learns a robust affine-invariant geometric prior with detail-faithful structure and sharp boundaries, which in turn makes the subsequent metric calibration both easier to optimize and more stable.

\section{Related Work} 

\label{sec:formatting}

\vspace{0.05in}\noindent \textbf{Relative depth estimation.} 
Recent progress in relative depth estimation~\cite{lyu2026stabilizing,wang2025vggt,dens3r,MoRE2025} has been largely driven by more expressive architectures and large-scale supervision. DPT~\cite{ranftl2021vision} leverages transformer backbones to more effectively capture global context, while Depth Anything~\cite{yang2024depth} and Depth Anything V2~\cite{yang2024depth2} scale training to millions of diverse images, achieving strong zero-shot generalization. Marigold~\cite{ke2025marigold} demonstrates that diffusion models contain strong geometric priors that can be adapted for depth prediction, and Moge~\cite{wang2025moge} improves structural consistency through explicit geometry-aware constraints. These advances have led to highly accurate and robust relative depth estimation.

\vspace{0.05in} \noindent \textbf{Metric depth estimation.} 
Metric depth estimation focuses on resolving the inherent scale ambiguity in monocular predictions. Metric3D~\cite{yin2023metric3d} and Metric3D v2~\cite{hu2024metric3d} address this by explicitly modeling camera intrinsics and incorporating geometric constraints, enabling more stable and accurate absolute depth recovery. ZoeDepth~\cite{bhat2023zoedepth} offers a practical pathway from relative to metric depth by using a shared encoder and lightweight domain-specific metric heads.

\vspace{0.05in}\noindent \textbf{Metric geometry estimation.} 
MoGe-2~\cite{wang2025moge2} improves single-image geometry by mitigating scale-shift ambiguity and enforcing global 3D consistency. DepthPro~\cite{bochkovskiydepth} leverages pretrained priors and coarse-to-fine structure with predicted focal length to recover fine-grained structures across diverse scenes. UniDepth~\cite{piccinelli2024unidepth} and UniDepth V2~\cite{piccinelli2025unidepthv2} unify depth and camera parameter estimation within a single framework, enabling camera-agnostic metric reconstruction without dataset-specific calibration.

\section{FoundationGeo}

\label{sec:methods} 

\begin{figure*}[t]
    \centering
    \includegraphics[width=1.0\linewidth]{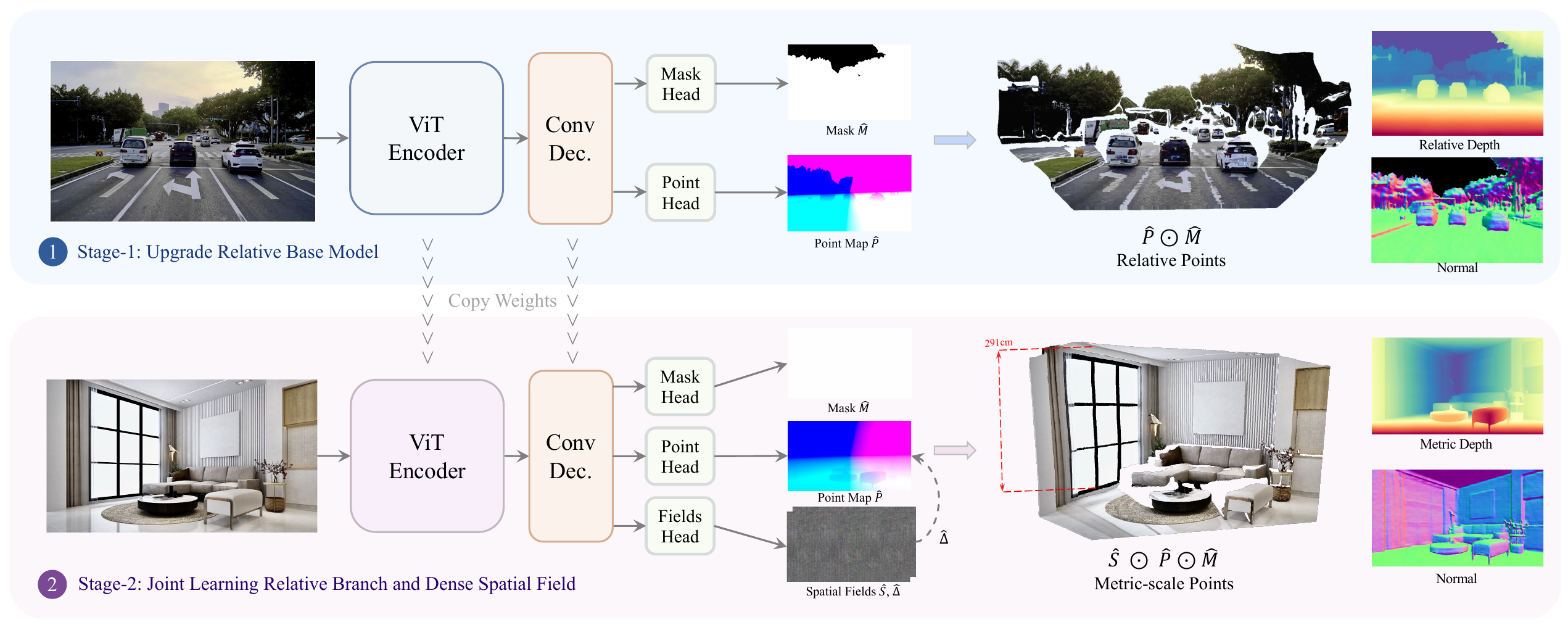}
    \caption{A ViT encoder with a lightweight up-sampling convolutional decoder first learns a high-fidelity {relative geometry} branch, predicting a validity mask $\hat{\mathbf M}$ and an affine-invariant point map $\hat{\mathbf P}$. In the second stage, we first apply a ray-direction correction field $\hat{\boldsymbol\Delta}$ to $\hat{\mathbf P}$ to obtain a direction-refined relative point map, and then use a spatial scale field $\hat{\mathbf S}$ to perform spatially varying rescaling, producing a metric point map $\tilde{\mathbf P}$. Metric depth and surface normals are subsequently derived from $\tilde{\mathbf P}$.}
    \label{fig:structure}
    \vspace{-12pt}
\end{figure*}

We introduce \textbf{FoundationGeo}, a two-stage framework that transforms affine-invariant geometry into physically grounded metric 3D understanding.
We first train a high-fidelity relative base model that predicts an affine-invariant point map with strong geometric structure (Sec.~\ref{3.1}).
Then we lift this relative geometry to metric space using spatial calibration fields: a learnable ray-direction correction field $\hat{\boldsymbol\Delta}$ refines the relative point-map rays to produce a direction-corrected affine-invariant point map $\hat{\mathbf P}$, and a learnable pixel-wise scale field $\hat{\mathbf S}\in\mathbb R^{H\times W}$ performs spatially varying metric calibration via $\tilde{\mathbf P}=\hat{\mathbf S}\odot\hat{\mathbf P}$, yielding metrically consistent 3D points $\tilde{\mathbf P}$ (Sec.~\ref{3.2}).
Finally,  we diagnose camera-model mismatch as a remaining bottleneck for zero-shot metric transfer and show that improving focal-length coverage through targeted rendered data helps reduce the residual relative to metric gap (Sec.~\ref{3.3}). An overview of our method is illustrated in Fig.~\ref{fig:structure}.
\vspace{-6pt}

\subsection{Upgrade Relative Base Model}

\label{3.1}

A high-quality affine-invariant relative base model is crucial for reliable metric calibration. We therefore begin by training a strong relative base model that is globally consistent and locally detailed.
As shown in Fig.~\ref{fig:structure} stage-1, we adopt a DINOv3-initialized~\cite{simeoni2025dinov3} ViT encoder to extract global, context-rich tokens, and attach a lightweight upsampling CNN decoder~\cite{wang2025moge} for dense regression of the affine-invariant point map $\hat{\mathbf P}$ and the reliability mask $\hat{\mathbf M}$.

The network is optimized with global and local supervision to capture large-scale layout and fine structures, the loss in this stage include $\mathcal{L}_\text{relative}$: (i) a global alignment loss that fits $\hat{\mathbf P}$ to ground-truth points under scale–shift ambiguity, (ii) multi-scale local patch losses at progressively finer resolutions to preserve edges and high-frequency details~\cite{wang2025moge}, (iii) a surface-normal consistency loss to encourage piecewise smooth yet detail-retaining surfaces~\cite{qi2018geonet,qi2020geonet++}, (iv) an edge loss, enabled only for datasets with abundant fine details and (v) a mask loss that trains $\hat{\mathbf M}$ to down-weight unreliable regions. We further leverage the encoder multi-scale feature maps by fusing them via element-wise summation. This lightweight fusion exposes the decoder to complementary cues at different semantic levels, which significantly enhances the fidelity of fine-grained details in our predicted relative geometry. 

To support robust generalization, we curate a 10.2M-image training corpus spanning indoor, outdoor, driving, synthetic, and in-the-wild domains (Table~\ref{datasets}). And we apply targeted data filtering to prioritize high-quality training data: (i) discard frames with extreme motion blur or over/under-exposure; (ii) exclude top-down/overhead views that lack near–far ordering; (iii) clip or invalidate depth beyond dataset-specific physical ranges. Overall, this stage yields a high-fidelity relative backbone that sets the upper bound for the subsequent spatial fields learning. Additional implementation details can be referred to \textit{suppl. materials}.

\subsection{Spatial Fields for Metric Geometry}

\label{3.2}
Prior relative-to-metric approaches typically rely on a single global scale factor~\cite{wang2025moge2}, which breaks down under spatially varying scale drift and ray-direction bias. We therefore introduce two lightweight spatial fields on top of the relative backbone: a ray-direction correction field $\hat{\boldsymbol\Delta}$ to refine the relative point map, and a pixel-wise scale field $\hat{\mathbf S}\in\mathbb{R}^{H\times W}$ for spatially varying metric calibration. 

\vspace{0.1in} \noindent \textbf{Ray direction correction.}  
Let $\hat{\mathbf p}_i\in\mathbb{R}^3$ be the predicted affine-invariant point at pixel $i$. We decompose it into a range term and a unit ray direction, with $\hat{d}_i=\|\hat{\mathbf p}_i\|_2$ and $\hat{\mathbf r}_i=\frac{\hat{\mathbf p}_i}{\|\hat{\mathbf p}_i\|_2}$.
We aim to correct $\hat{\mathbf r}_i$ while preserving $\hat{d}_i$, so that the correction is scale-invariant and does not interfere with subsequent metric scaling. To this end, we construct two orthonormal tangent directions $(\mathbf b_{1,i},\mathbf b_{2,i})$ spanning the plane orthogonal to $\hat{\mathbf r}_i$ (i.e., $\mathbf b_{1,i}\perp \hat{\mathbf r}_i$, $\mathbf b_{2,i}\perp \hat{\mathbf r}_i$, and $\mathbf b_{1,i}\perp \mathbf b_{2,i}$). In practice, we robustly build the tangent basis by selecting a reference axis that is not nearly parallel to $\hat{\mathbf r}_i$, and obtain $\mathbf b_{1,i},\mathbf b_{2,i}$ via cross products followed by normalization (details in the supplementary material).

The ray field head predicts $\hat{\boldsymbol{\Delta}}_i=(\hat{\Delta}_{1,i},\hat{\Delta}_{2,i})$, which parameterizes a bounded angular perturbation along the tangent basis, where $\delta_{1,i}=\delta_{\max}\tanh(\hat{\Delta}_{1,i})$ and $\delta_{2,i}=\delta_{\max}\tanh(\hat{\Delta}_{2,i})$, with $\delta_{\max}$ being a small bound that stabilizes training. We then update the ray direction as $\hat{\mathbf r}'_i=\frac{\hat{\mathbf r}_i+\delta_{1,i}\mathbf b_{1,i}+\delta_{2,i}\mathbf b_{2,i}}{\left\|\hat{\mathbf r}_i+\delta_{1,i}\mathbf b_{1,i}+\delta_{2,i}\mathbf b_{2,i}\right\|_2}$, and reconstruct the direction-corrected relative point by restoring the original range, $\hat{\mathbf p}'_i=\hat{d}_i\hat{\mathbf r}'_i$.

\vspace{0.1in}\noindent \textbf{Closed-form metric calibration.}
Given the direction-corrected affine-invariant point prediction $\hat{\mathbf p}^{\,\prime}_i\in\mathbb R^3$, we recover metric geometry by predicting the inherent scaling factor. As demonstrated by Fig.~\ref{fig:Observation}, a single global scale usually leads to suboptimal results; we instead propose to predict pixel-wise scales to better account for the spatial variation in the predicted point maps. Specifically, we use a learnable output block, which takes shared pointmap branch features as input, to predict a pixel-wise scale field $\hat{s}_i$, and the final metric geometry is obtained via applying $\tilde{\mathbf p}_i = \hat{\mathbf p}^{\,\prime}_i \cdot \hat{s}_i$.

\vspace{0.1in}\noindent \textbf{Training objectives.}
To jointly optimize the direction-refined point prediction $\hat{\mathbf p}^{\,\prime}_i$ and the spatial fields prediction $\hat{s}_i$, $\hat{\boldsymbol\Delta}_i$ we use a coupled metric l1 loss:
\begin{equation}
{\small
\mathcal L_{\rm metric}
= \sum_{i\in\mathcal M} \frac{1}{z_i}\,\bigl\|\tilde{\mathbf p}_i-\mathbf p_i\bigr\|_1,
}
\label{eq:global_loss}
\end{equation}
where $\tilde{\mathbf p}_i=\hat{\mathbf p}^{\,\prime}_i\cdot \hat{s}_i$ is the predicted metric point, $\mathbf p_i$ is the ground-truth 3D point, $\mathcal M$ is the set of valid pixels with metric supervision, and $z_i$ is the depth of the ground-truth point used to up-weight near-range accuracy.

We define the normalized weighted average over valid pixels as
$\langle f_i\rangle_{\mathcal M}=\frac{\sum_{i\in\mathcal M} w_i f_i}{\sum_{i\in\mathcal M} w_i}$,
with $w_i=1/z_i$.
We supervise the ray correction by enforcing angular consistency between the predicted metric rays and ground-truth rays:
\begin{equation}
{\small
\mathcal L_{\rm ray}
= \Big\langle \, \bigl|\angle(\tilde{\mathbf r}_i, \mathbf r_i)\bigr|_{\beta_{\rm ray}} \, \Big\rangle_{\mathcal M},
}
\label{eq:ray_loss}
\end{equation}
where $\tilde{\mathbf r}_i=\tilde{\mathbf p}_i/\|\tilde{\mathbf p}_i\|_2$ and $\mathbf r_i=\mathbf p_i/\|\mathbf p_i\|_2$ are unit ray directions, $\angle(\cdot,\cdot)$ denotes the angular difference, and $|\cdot|_{\beta_{\rm ray}}$ is a Huber penalty with threshold $\beta_{\rm ray}$.
To prevent unnecessary ray drift, we regularize the magnitude of the bounded correction.
Recall that the applied angular offsets are bounded by $\delta_{k,i}=\delta_{\max}\tanh(\hat\Delta_{k,i})$.
We penalize the bounded correction as
\begin{equation}
{\small
\mathcal L_{\Delta}
= \Big\langle \, |\tanh(\hat\Delta_{1,i})|^{q} + |\tanh(\hat\Delta_{2,i})|^{q} \, \Big\rangle_{\mathcal M}.
}
\label{eq:delta_reg}
\end{equation}
where $\hat{\boldsymbol\Delta}_i=(\hat\Delta_{1,i},\hat\Delta_{2,i})$ is the predicted correction field.
In all experiments we set $q=2$, which encourages small corrections unless supported by supervision.

While end-to-end metric supervision can jointly learn $\hat{\mathbf p}^{\,\prime}_i$ and $\hat{s}_i$, it may entangle their roles and blur the distinction between geometry prediction and scaling correction. To provide a direct and stable supervision signal for the scale field, we compute a closed-form target scale $s_i$ by projecting the ground-truth point $\mathbf p_i$ onto the direction-refined affine-invariant prediction $\hat{\mathbf p}^{\,\prime}_i$, with
$s_i = \frac{(\hat{\mathbf p}^{\,\prime}_i)^\top \mathbf p_i}{\|\hat{\mathbf p}^{\,\prime}_i\|_2^2}$.
This is the least-squares minimizer of $\|\hat{\mathbf p}^{\,\prime}_i\cdot s - \mathbf p_i\|_2^2$ with respect to the scalar $s$. 
To directly supervise the scale field and decouple it from the point-map head, we define the scale-field loss as
\begin{equation}
{\small
\mathcal L_{\rm scalefield}
= \Big\langle \, \bigl|\log \hat{s}_i - \log s_i^{\ast}\bigr|_\beta \, \Big\rangle_{\mathcal M},
}
\label{eq:scalefield_loss}
\end{equation}
where $\langle\cdot\rangle_{\mathcal M}$ denotes the normalized weighted average over valid pixels and $|x|_\beta$ is the Huber penalty with threshold $\beta$. We clamp the target scale into a bounded physical range and supervise in the log domain, with $\log s_i^{\ast} = \log\!\Bigl(\mathrm{clamp}(s_i,\, s_{\min},\, s_{\max})\Bigr)$.

Overall, we optimize FoundationGeo with a unified objective that combines the relative-geometry supervision from Stage~I and the metric calibration losses from Stage~II:
\begin{equation}
{\small
\mathcal L_{\rm FoundationGeo}
=
\mathcal L_{\rm relative}
+
\mathcal L_{\rm metric}
+
\gamma_{\rm s}\mathcal L_{\rm scalefield}
+
\gamma_{\rm r}\mathcal L_{\rm ray}
+
\gamma_{\Delta}\mathcal L_{\Delta},
}
\label{eq:overall_obj}
\end{equation}
where $\mathcal L_{\rm metric}$ is the coupled metric regression loss in Eq.~(\ref{eq:global_loss}), and $\mathcal L_{\rm scalefield}$, $\mathcal L_{\rm ray}$, and $\mathcal L_{\Delta}$ are the decoupled spatial-field supervision terms defined in Eqs.~(\ref{eq:scalefield_loss}), (\ref{eq:ray_loss}), and (\ref{eq:delta_reg}), respectively. 
The latter three losses are weighted by $\gamma_{\rm s}$, $\gamma_{\rm r}$, and $\gamma_{\Delta}$ to ensure that the scale field and ray correction field act as lightweight calibration modules rather than redundantly absorbing geometric prediction. 
We keep $\mathcal L_{\rm relative}$ active throughout training so that samples still contribute to robust representation learning and generalization. 
The detailed weighting strategy and hyperparameter settings are provided in the supplementary material.

\subsection{Focal-Length Coverage Analysis and Synthetic Augmentation}
\label{3.3}  
Although our spatial fields effectively calibrate relative predictions into metric geometry (Sec.~\ref{3.1}--\ref{3.2}), we still observe a residual gap in zero-shot transfer, suggesting that metric accuracy is additionally limited by training-data coverage and camera-model diversity. We therefore perform diagnostic studies on distribution mismatch in camera intrinsics, with a focus on focal-length statistics across seven evaluation benchmarks.

\begin{figure}[h]
    \centering
    \vspace{-12pt}
    \includegraphics[width=1\linewidth]{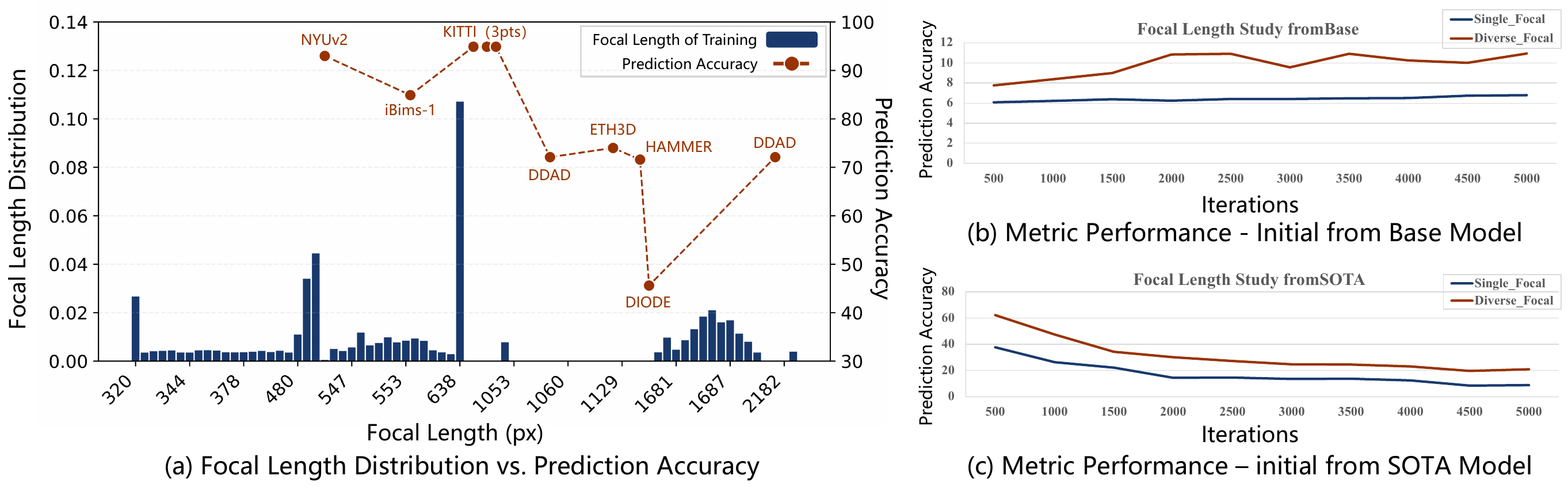}
    \caption{(a) Training focal distribution (top-50 frequent values) vs.\ benchmark performance.
(b)(c) Controlled Blender fine-tuning with \textit{Single-Focal} vs.\ \textit{Diverse-Focal} for (b) our base model and (c) a pre-trained metric model.}
    \vspace{-18pt}
    \label{fig:focal_study}
\end{figure}

As illustrated in Fig.~\ref{fig:focal_study}(a), monocular metric prediction is fundamentally coupled with focal length~\cite{hu2024metric3d,yin2023metric3d}. When training covers only a limited set of camera models, the network may internalize a biased implicit focal prior, leading to systematic over- or under-scaling on unseen optics. To quantify this effect, we analyze and visualize the top-50 most frequent focal values in our 10.2M-image training corpus and compare them with those of the zero-shot benchmarks.

Interestingly, we observe a clear correlation between distribution overlap and metric accuracy. Datasets whose focal lengths closely align with our training distribution (e.g., NYUv2, KITTI, iBims-1) show strong accuracy with minimal scale drift, whereas camera-mismatched benchmarks (e.g., DDAD, ETH3D, HAMMER, DIODE) concentrate in a narrow focal band (approximately 1000--1373) that is under-covered by our training data and exhibit degraded performance. These results confirm that the remaining performance gap primarily stems from out-of-distribution focal lengths, where the model’s implicit scale prior becomes unreliable.

Rather than explicitly feeding focal length or predicting it as an auxiliary variable~\cite{piccinelli2024unidepth,piccinelli2025unidepthv2,bochkovskiydepth}, we explore a purely data-driven strategy that expands focal coverage so the model can learn the image--depth--camera coupling directly from paired supervision. Concretely, we render two matched synthetic datasets with identical scenes and camera trajectories: \textit{Single-Focal}, rendered with a fixed focal length, and \textit{Diverse-Focal}, rendered with multiple focal values spanning a wide range. Fine-tuning both our Base Model and a pre-trained metric model under identical recipes shows that \textit{Diverse-Focal} leads to consistently stronger cross-dataset zero-shot behavior than \textit{Single-Focal} (Fig.~\ref{fig:focal_study}(b,c)), supporting focal diversity as a practical lever for robustness.

Guided by this diagnostic, we use a Blender-based data engine to synthesize \textit{23,700} additional training images across \textit{7} diverse indoor and outdoor scenes (Fig.~\ref{fig:vis}), targeting the under-covered focal regime. Rather than merely scaling up data, this set serves as a controlled intervention on camera intrinsics, encouraging the model to better capture the depth--focal coupling under pointmap supervision. Concretely, we render varying  focal lengths that uniformly span the missing band and inject these samples into the metric training stage. This targeted coverage helps mitigate the biased implicit focal prior that can lead to systematic over- or under-scaling on unseen optics. Importantly, the effect is complementary to our spatial calibration fields: the scale and ray-direction fields address within-image spatial drift and directional bias, while targeted focal coverage improves across-camera generalization by aligning the model's implicit intrinsic prior. As a result, metric predictions become less sensitive to intrinsic shifts, further reducing the residual gap on camera-mismatched benchmarks (details in the supplementary material).

\section{Experiments}
\label{sec:experiments}

\vspace{0.05in} \noindent \textbf{Implementation Details.} Our FoundationGeo model is trained in two stages. In the first stage, we train an improved relative depth estimation model with an advanced ViT-Large~\cite{dosovitskiy2021imageworth16x16words} encoder pre-trained with DINOv3~\cite{simeoni2025dinov3}. During training, the encoder and decoder are trained with initial learning rates of $1\times 10^{-5}$ and $1\times 10^{-4}$, respectively, and the learning rate is halved every 20K iterations. 
In the second stage, we further fine-tune the relative depth estimation model to yield the metric depth. Specifically, we set $\gamma_{\rm s}$, $\gamma_{\rm r}$, and $\gamma_{\Delta}$ as 0.2, 0.1 0.05; and fine-tune the ViT backbone and decoder heads with learning rates of $1\times 10^{-6}$ and $1\times 10^{-5}$, respectively. To ensure robustness, image augmentations, including color jittering, Gaussian blurring, JPEG compression-decompression, and random cropping, are employed in both training stages. The full model is trained for 55K (first stage) + 20K (second stage) iterations using 32 NVIDIA H20 GPUs.

\vspace{0.05in} \noindent \textbf{Datasets.} For training, as shown in Table~\ref{datasets}, we train our model on 19 datasets with a total number of 10.2 million frames. These training datasets span a wide range of scenarios and cover heterogeneous GT sources (e.g., synthetic renderings, LiDAR, RGB-D sensors, and SfM reconstructions). Moreover, to obtain high-quality training data, we perform data filtering as described in Sec.~\ref{3.1}. For testing, we evaluate the accuracy of depth estimation on 8 datasets (i.e., NYUv2~\cite{silberman2012indoor}, KITTI~\cite{Uhrig2017THREEDV}, ETH3D~\cite{schops2019bad}, iBims-1~\cite{koch2018evaluation,koch2020comparison}, Sintel~\cite{butler2012naturalistic}, DDAD~\cite{guizilini20203d}, DIODE~\cite{vasiljevic2019diode} and HAMMER~\cite{jung2023importance}), which are excluded from training, to demonstrate the superiority of our method and the zero-shot generalization capability.

\vspace{0.05in} \noindent \textbf{Evaluation Metrics.} We measure the accuracy (\textit{AbsRel} and $\delta_1$) and quality (boundary F1 ) of the estimated depth.
\textit{AbsRel} reflects the magnitude of scale-sensitive depth errors, while $\delta_1$ measures inlier consistency as the fraction of pixels within a relative-error threshold.
For metric depth, we report the absolute relative error
\textit{AbsRel}, i.e.,
$|\tilde d-d|/d$, and the percentage of inliers $\delta_1$ satisfying
$\max(d/\tilde d,\tilde d/d)<1.25$.
For relative depth, we report affine-invariant \textit{AbsRel}, i.e.,
$|\hat d-d|/d$, and the percentage of inliers $\delta_1$ satisfying
$\max(d/\hat d,\hat d/d)<1.25$.
For boundary sharpness, we report boundary F1 following Depth Pro~\cite{bochkovskiydepth}.

\vspace{0.05in} \noindent \textbf{Baselines.}
We mainly follow the evaluation protocol of MoGe-2~\cite{wang2025moge2} unless otherwise specified. \textit{1) Metric depth comparisons.} We evaluate the official metric variants of Depth Anything v1~\cite{yang2024depth} and v2~\cite{yang2024depth2}. Since these models are released as separate indoor/outdoor variants, we run both variants on all benchmarks and report the better-performing result as the baseline score. For Metric3D v2~\cite{hu2024metric3d}, the model requires ground-truth camera intrinsics as input; we therefore include it for reference but explicitly annotate it and exclude it from the aggregated ranking. All other baselines provide native metric-depth outputs, and we directly use their predictions under the corresponding evaluation protocol. \textit{2) Relative depth comparisons.} In addition to methods that natively predict relative depth, we also report relative-depth results derived from metric models. Concretely, we align each method's predicted depth to the ground truth using a single global scale and shift before computing the relative-depth scores to ensure a consistent comparison. In this setting, we evaluate Depth Anything v3~\cite{lin2025depth} using its official {monocular} variant. We additionally include the single-frame performance of the multi-view method VGGT~\cite{wang2025vggt}.

\vspace{-8pt}

\subsection{Main Results}

\begin{table*}[!t]

    \caption{Summary of datasets used for training.}

    \label{datasets}
    \centering
    \renewcommand{\arraystretch}{1.2}
    \setlength{\tabcolsep}{5pt}
    \footnotesize

    \begin{minipage}[t]{0.49\textwidth}
        \centering
        \resizebox{\linewidth}{!}{
        \begin{tabular}{l c c c c}
            \specialrule{0.12em}{0em}{0em}
            Name & Domain & \# Frames & Syn. & Metric \\
            \hline
            Argoverse2~\cite{Argoverse2} & Outdoor/Driving & $1.5$M  & N & \ding{51}\\
            ARKitScenes~\cite{baruch1arkitscenes} & Indoor & $441$K & N & \ding{51}\\
            BlendedMVS~\cite{yao2020blendedmvs} & In-the-wild & $109$K & N & \ding{51}\\
            Taskonomy~\cite{zamir2018taskonomy} & Indoor & $4.6$M & N & \ding{51}\\
            Waymo~\cite{sun2020scalability} & Outdoor/Driving & $790$K & N & \ding{51}\\
            Voyager & Outdoor/Driving & $221$K & N & \ding{51}\\
            FSD~\cite{wen2025stereo} & Outdoor/In-the-wild & $1.04$M & Y & \\
            Hypersim~\cite{roberts2021hypersim} & Indoor & $64$K & Y & \ding{51}\\
            IRS~\cite{wang2019irs} & Indoor & $94$K & Y & \ding{51}\\
            KenBurns~\cite{niklaus20193d} & In-the-wild & $72$K & Y & \\
            \specialrule{0.12em}{0em}{0em}
        \end{tabular}}
    \end{minipage}\hfill
    \begin{minipage}[t]{0.49\textwidth}
        \centering
        \resizebox{\linewidth}{!}{
        \begin{tabular}{l c c c c}
            \specialrule{0.12em}{0em}{0em}
            Name & Domain & \# Frames & Syn. & Metric \\
            \hline
            MatrixCity~\cite{li2023matrixcity} & Outdoor/Driving & $354$K & Y & \ding{51}\\
            MidAir~\cite{fonder2019mid} & Outdoor/In-the-wild & $423$K & Y & \ding{51}\\
            MVS-Synth~\cite{huang2018deepmvs} & Outdoor/Driving & $12$K & Y & \ding{51}\\
            Spring~\cite{mehl2023spring} & In-the-wild & $5$K & Y & \\
            Structured3D~\cite{zheng2020structured3d} & Indoor & $76$K & Y & \\
            TartanAir~\cite{wang2020tartanair} & In-the-wild & $259$K & Y & \ding{51}\\
            UrbanSyn~\cite{gomez2025all} & Outdoor/Driving & $7$K & Y & \ding{51}\\
            Dynamic-Replica~\cite{karaev2023dynamicstereo} & Indoor & $143$K & Y & \ding{51}\\
            FoundationGeo & Indoor/Outdoor & $23$K & Y & \ding{51}\\
            \hline
            \textbf{Total} & & \textbf{10.2M} & & \\
            \specialrule{0.12em}{0em}{0em}
        \end{tabular}}
    \end{minipage}
\vspace{-6pt}

\end{table*}

We assess the zero-shot performance of our framework and compare it to several state-of-the-art methods on monocular metric depth estimation, affine-invariant depth estimation and boundary sharpness.

\vspace{0.04in}\noindent \textbf{Quantitative results for metric depth estimation:} 
Table~\ref{table:benchmark} shows that, although the best method varies by dataset, {FoundationGeo is the most balanced overall}, achieving the lowest average \textit{AbsRel} and highest average $\delta_1$, and thus the best average \textit{Rank} (2.32) across the seven benchmarks. The key takeaway is robustness to domain and camera-model shifts: some methods excel in a narrow regime but degrade sharply on specific domains. For example, UniDepth nearly fails on ETH3D (56.9 \textit{AbsRel}, 14.9 $\delta_1$), and MoGe-2 drops substantially on KITTI (18.1 \textit{AbsRel}, 62.9 $\delta_1$), whereas FoundationGeo remains consistently competitive across both indoor/driving and challenging datasets. We attribute this stability to our data and training design: a stronger relative depth estimation backbone with broader camera-model coverage, combined with the spatial calibration fields to ensure accuracy. Notably, FoundationGeo achieves the best performance on HAMMER (22.4 \textit{AbsRel}, 69.6 $\delta_1$), an object-centric benchmark with complex scene composition.

\begin{table*}[t]
  \caption{Quantitative results for metric and relative depth estimation. \textit{AbsRel} and $\delta_1$ are in percentage. The best values are highlighted in \textbf{bold}, and the second-best ones are \underline{underlined}. * means model needs GT intrinsic as input. \textcolor{gray!50}{Gray numbers} denote models trained on respective benchmarks or need GT intrinsics, thus excluded from ranking.}
  \label{table:benchmark}
  \centering
  \setlength{\tabcolsep}{2.0pt}
  \renewcommand{\arraystretch}{1.35}
  \scriptsize
  \resizebox{\textwidth}{!}{%
  \begin{tabular}{l|cc|cc|cc|cc|cc|cc|cc|cc|ccc}
    \toprule
    \multirow{2}{*}{Method}
      & \multicolumn{2}{c|}{NYUv2}
      & \multicolumn{2}{c|}{KITTI}
      & \multicolumn{2}{c|}{ETH3D}
      & \multicolumn{2}{c|}{iBims-1}
      & \multicolumn{2}{c|}{Sintel}
      & \multicolumn{2}{c|}{DDAD}
      & \multicolumn{2}{c|}{DIODE}
      & \multicolumn{2}{c|}{HAMMER}
      & \multicolumn{3}{c}{Average} \\
      & AbsRel$\downarrow$ & $\delta_1\uparrow$
      & AbsRel$\downarrow$ & $\delta_1\uparrow$
      & AbsRel$\downarrow$ & $\delta_1\uparrow$
      & AbsRel$\downarrow$ & $\delta_1\uparrow$
      & AbsRel$\downarrow$ & $\delta_1\uparrow$
      & AbsRel$\downarrow$ & $\delta_1\uparrow$
      & AbsRel$\downarrow$ & $\delta_1\uparrow$
      & AbsRel$\downarrow$ & $\delta_1\uparrow$
      & AbsRel$\downarrow$ & $\delta_1\uparrow$ & Rank$\downarrow$ \\
    \midrule

    \multicolumn{20}{c}{\textbf{Metric Depth Estimation}} \\
    \midrule
    ZoeDepth~\cite{bhat2023zoedepth}
      & \textcolor{gray!50}{11.0} & \textcolor{gray!50}{91.9}
      & \textcolor{gray!50}{17.0} & \textcolor{gray!50}{85.4}
      & 57.1 & 33.7
      & 17.4 & 67.2
      & -- & --
      & 38.9 & 38.6
      & 39.3 & 29.3
      & 94.3 & 3.23 
      & \textcolor{gray!50}{39.3} & \textcolor{gray!50}{49.9} & 6.90 \\
    MASt3R~\cite{leroy2024grounding}
      & 10.8 & 89.7
      & 56.7 & 9.84
      & 47.2 & 20.1
      & 18.7 & 61.5
      & -- & --
      & 62.4 & 5.51
      & 54.9 & 19.0
      & 97.2 & 6.74
      & 49.7 & 30.3 & 7.71 \\
    DA V1~\cite{yang2024depth}
      & \textcolor{gray!50}{10.5} & \textcolor{gray!50}{94.9}
      & \textcolor{gray!50}{11.6} & \textcolor{gray!50}{94.5}
      & 40.2 & 24.0
      & 12.9 & 81.8
      & -- & --
      & 34.5 & 44.7
      & 58.0 & 16.2
      & 54.8 & 27.3 
      & \textcolor{gray!50}{31.8} & \textcolor{gray!50}{54.8} & 6.50 \\
    DA V2~\cite{yang2024depth2}
      & 16.4 & 80.9
      & 10.6 & 88.6
      & 36.1 & 36.3
      & 11.1 & \underline{91.7}
      & -- & --
      & 41.7 & 37.5
      & 41.2 & 22.1
      & 52.1 & 38.9
      & 29.9 & 56.6 & 5.36\\
    UniDepth V1~\cite{piccinelli2024unidepth}
      & \underline{7.59} & \textbf{97.6}
      & \textbf{4.69} & \textbf{98.4}
      & 56.9 & 14.9
      & 23.8 & 57.6
      & -- & --
      & \textbf{13.8} & \textbf{85.1}
      & \textbf{17.1} & \textbf{71.9}
      & 38.2 & 46.7
      & 23.2 & 67.5 & 3.75\\
    UniDepth V2~\cite{piccinelli2025unidepthv2}
      & 10.6 & 92.8
      & 8.58 & \underline{95.4}
      & 20.7 & 69.5
      & \textbf{9.52} & \textbf{93.2}
      & -- & --
      & 18.4 & \underline{77.6}
      & 43.0 & 51.8
      & 38.2 & 46.8
      & 21.3 & 75.3 & 3.25\\
    DepthPro~\cite{bochkovskiydepth}
      & 10.7 & 91.9
      & 23.5 & 38.3
      & 38.5 & 32.8
      & 15.9 & 81.5
      & -- & --
      & 33.4 & 35.3
      & 31.9 & 37.7
      & 39.1 & 63.0
      & 27.6 & 54.4 & 5.36 \\
    Metric3D V2*~\cite{hu2024metric3d}
      & \textcolor{gray!50}{7.16} & \textcolor{gray!50}{96.5}
      & \textcolor{gray!50}{5.25} & \textcolor{gray!50}{98.0}
      & \textcolor{gray!50}{11.8} & \textcolor{gray!50}{88.8}
      & \textcolor{gray!50}{9.96} & \textcolor{gray!50}{94.1}
      & -- & --
      & \textcolor{gray!50}{9.21} & \textcolor{gray!50}{93.7}
      & \textcolor{gray!50}{49.1} & \textcolor{gray!50}{1.98}
      & \textcolor{gray!50}{35.7} & \textcolor{gray!50}{44.3}
      & \textcolor{gray!50}{18.3} & \textcolor{gray!50}{73.9} & - \\
    MoGe-2~\cite{wang2025moge2}
      & \textbf{7.33} & \underline{96.1}
      & 18.1 & 62.9
      & \textbf{10.4} & \textbf{90.8}
      & 13.6 & 83.0
      & -- & --
      & \underline{15.8} & 73.0
      & \underline{17.5} & {66.4}
      & \underline{26.9} & \underline{65.6}
      & \underline{15.7} & \underline{76.8} & \underline{2.82}\\
    \rowcolor{gray!12}
    {FoundationGeo}
      & 10.2 & 93.0
      & \underline{7.86} & 94.9
      & \underline{17.8} & \underline{74.0}
      & \underline{10.4} & 90.0
      & -- & --
      & 17.6 & 74.7
      & \underline{17.5} & \underline{69.7}
      & \textbf{22.4} & \textbf{69.6}
      & \textbf{14.8} & \textbf{80.8} & \textbf{2.32}\\

    \midrule
    \multicolumn{20}{c}{\textbf{Relative Depth Estimation}} \\
    \midrule
    ZoeDepth~\cite{bhat2023zoedepth}
      & \textcolor{gray!50}{4.76} & \textcolor{gray!50}{97.3}
      & \textcolor{gray!50}{5.59} & \textcolor{gray!50}{95.1}
      & 7.27 & 94.2
      & 5.85 & 95.7
      & 21.8 & 69.2
      & 14.2 & 80.1
      & 7.80 & 90.9
      & 6.65 & 95.7 
      & \textcolor{gray!50}{9.24} & \textcolor{gray!50}{89.8} & 13.08\\
    PPD~\cite{xupixel}
      & 5.16 & 97.0
      & 11.5 & 83.0
      & 7.98 & 91.1
      & 5.34 & 95.9
      & 18.8 & 74.4
      & 18.8 & 70.3
      & 7.68 & 90.5
      & 3.77 & 99.2
      & 9.88 & 87.7 & 11.66 \\
    MASt3R~\cite{leroy2024grounding}
      & 4.67 & 96.7
      & 5.79 & 95.1
      & 4.64 & 97.0
      & 4.62 & 95.6
      & 21.3 & 70.3
      & 12.5 & 83.4
      & 5.79 & 94.1
      & 4.21 & 96.8
      & 7.94 & 91.1 & 10.78 \\
    VGGT~\cite{wang2025vggt}
      & 3.01 & 98.4
      & 5.34 & 95.1
      & 3.47 & 97.7
      & 3.64 & 96.9
      & 18.1 & 76.2
      & 14.0 & 81.1
      & 5.15 & 94.6
      & 3.60 & 97.5
      & 7.04 & 92.2 & 8.59 \\
    DA V1~\cite{yang2024depth}
      & \textcolor{gray!50}{3.82} & \textcolor{gray!50}{98.3}
      & \textcolor{gray!50}{5.04} & \textcolor{gray!50}{96.4}
      & 6.23 & 95.2
      & 4.23 & 97.3
      & 20.1 & 71.8
      & 11.3 & 86.1
      & 6.75 & 92.6
      & 5.77 & 97.3
      & \textcolor{gray!50}{7.91} & \textcolor{gray!50}{91.9} & 10.58 \\
    DA V2~\cite{yang2024depth2}
      & 4.16 & 97.9
      & 6.77 & 94.3
      & 4.63 & 97.2
      & 3.44 & 98.3
      & 17.1 & 76.6
      & 13.4 & 81.8
      & 5.41 & 94.6
      & 4.73 & 98.9
      & 7.46 & 92.4 & 8.94 \\
    DA V3~\cite{lin2025depth}
      & 3.39 & 98.4
      & 4.80 & 97.1
      & 4.37 & 96.9
      & 2.96 & 98.6
      & 17.4 & 75.8
      & 11.4 & 86.7
      & 4.85 & 95.5
      & 3.30 & 99.4
      & 6.56 & 93.6 & 6.50 \\
    Metric3D V2~\cite{hu2024metric3d}
      & 3.94 & 97.6
      & \textbf{3.50} & \underline{98.4}
      & 3.24 & 99.0
      & 3.28 & 98.3
      & 26.6 & 71.7
      & \textcolor{gray!50}{7.15} & \textcolor{gray!50}{94.8}
      & \textbf{2.75} & \textbf{98.7}
      & 3.02 & 99.0
      & \textcolor{gray!50}{6.69} & \textcolor{gray!50}{94.7} & 5.86 \\
    UniDepth V1~\cite{piccinelli2024unidepth}
      & 3.40 & 98.6
      & \underline{3.55} & \textbf{98.7}
      & 4.92 & 97.5
      & 3.76 & 98.2
      & 24.9 & 64.1
      & 9.46 & 90.8
      & 4.90 & 96.2
      & 3.55 & 98.9
      & 7.31 & 92.9 & 7.19 \\
    UniDepth V2~\cite{piccinelli2025unidepthv2}
      & 2.96 & 98.6
      & 3.85 & 98.1
      & 2.95 & 98.5
      & \underline{2.64} & 98.4
      & 13.3 & 83.2
      & 10.5 & 90.9
      & 4.05 & 96.5
      & \textbf{2.48} & \textbf{99.6}
      & 5.34 & 95.5 & 3.50 \\
    DepthPro~\cite{bochkovskiydepth}
      & 3.67 & 98.2
      & 5.12 & 96.8
      & 4.97 & 96.4
      & 3.23 & 98.3
      & 15.8 & 80.1
      & 12.6 & 84.1
      & 4.66 & 95.6
      & 3.30 & \textbf{99.6}
      & 6.67 & 93.6 & 7.12 \\
    MoGe-1~\cite{wang2025moge}
      & \underline{2.92} & 98.6
      & 3.94 & 98.0
      & \textbf{2.69} & \underline{99.2}
      & 2.74 & 97.9
      & \underline{13.0} & \underline{83.2}
      & 8.40 & 92.1
      & 3.16 & \underline{97.5}
      & 3.00 & 98.3
      & 4.98 & 95.6 & 3.91 \\
    MoGe-2~\cite{wang2025moge2}
      & \textbf{2.89} & 98.6
      & 3.75 & 98.1
      & \underline{2.80} & 98.1
      & \textbf{2.36} & \underline{98.8}
      & 13.3 & 82.5
      & 8.26 & 92.5
      & \underline{3.14} & 97.4
      & 2.85 & 99.3
      & \underline{4.92} & \underline{95.7} & \underline{2.88} \\
    \rowcolor{gray!12}
    {FoundationGeo-Base}
      & 2.97 & \underline{98.7}
      & 3.96 & 98.1
      & 2.85 & \textbf{99.3}
      & 2.66 & \textbf{98.9}
      & \textbf{12.7} & \textbf{83.3}
      & \textbf{7.73} & \textbf{92.9}
      & 3.43 & \underline{97.5}
      & \underline{2.65} & 99.2
      & \textbf{4.87} & \textbf{96.0} & \textbf{2.56} \\
    \bottomrule
  \end{tabular}%
  }
  \vspace{-16pt}
\end{table*}

\vspace{0.04in} \noindent \textbf{Quantitative results for relative depth estimation.}
Although our focus is metric-scale geometry, we also evaluate affine-invariant depth to assess how effectively our finetuning improves scale-free predictions. Table~\ref{table:benchmark} shows that our \textit{FoundationGeo-Base} achieves the best overall relative performance across eight benchmarks, with the lowest average \textit{AbsRel}, the highest average $\delta_1$, and the best average \textit{Rank}. Beyond the average, the gains are most evident on cross-domain benchmarks: our base model leads on ETH3D (2.85 \textit{AbsRel}, 99.3 $\delta_1$), iBims-1 (2.66, 98.9), and Sintel (12.7, 83.3), and also performs strongly on driving-scale DDAD (7.73, 92.9), indicating a more transferable affine-invariant geometric prior. We attribute this to our upgraded training recipe: DINOv3 initialization, large-scale multi-domain training, detailed losses design, and multi-scale feature fusion, which improves fine-detail fidelity and reduces domain-specific failure modes, providing a stronger foundation for subsequent metric calibration.

\vspace{0.04in}\noindent\textbf{Quantitative measurements for boundary sharpness.}
We evaluate boundary sharpness using boundary F1~\cite{bochkovskiydepth} on Sintel~\cite{butler2012naturalistic}, iBims-1~\cite{koch2020comparison,koch2018evaluation}, and HAMMER~\cite{jung2023importance}. As shown in Table~\ref{tab:qualitative_comparison_boundary}, our Base Model is already comparable to Depth Pro with a 3.6k-token input, and increasing the token budget to 6k consistently improves boundary F1, achieving the best overall sharpness among compared methods despite Depth Pro operating at a higher native resolution (1536$\times$1536). Importantly, compared to the MoGe-style baseline we follow, our Base Model produces markedly sharper boundaries, which we attribute to our relative-branch upgrades including multi-scale feature fusion and detail-aware edge supervision that strengthen high-frequency geometric fidelity.

\begin{table*}[t]
    \vspace{-3mm}
    \centering
    \scriptsize
    \renewcommand{\arraystretch}{1.08}
    \caption{Evaluation of boundary sharpness using F1 scores ($\uparrow$) in percentages.}
    \label{tab:qualitative_comparison_boundary}
    \vspace{-4pt}

    \resizebox{0.68\textwidth}{!}{
    \begin{tabular}{l c ccc c}
    \toprule
    \textbf{Method} 
    & \#Parameters
    & iBims-1
    & HAMMER
    & Sintel
    & \textit{Avg. Rk.$\downarrow$} \\
    \midrule
    ZoeDepth~\cite{bhat2023zoedepth}          & 345M & 2.47 & 0.17 & 2.30 & 9.00 \\
    DA V1~\cite{yang2024depth}                & 335M & 3.68 & 0.76 & 5.64 & 8.00 \\
    DA V2~\cite{yang2024depth2}               & 335M & 13.9 & 4.74 & 32.5 & 4.33 \\
    Metric3D V2~\cite{hu2024metric3d}         & 412M & 7.36 & 1.40 & 25.3 & 7.00 \\
    MASt3R~\cite{leroy2024grounding}             & $\sim$700M & 1.24 & 0.05 & 1.72 & 10.67 \\
    UniDepth V1~\cite{piccinelli2024unidepth} & 347M & 2.35 & 0.06 & 0.73 & 10.33 \\
    UniDepth V2~\cite{piccinelli2025unidepthv2} & 354M & 11.2 & 4.40 & \underline{39.7} & 4.33 \\
    Depth Pro~\cite{bochkovskiydepth}         & 504M & 14.3 & \underline{5.36} & \textbf{41.6} & \underline{2.00} \\
    MoGe~\cite{wang2025moge}                  & 314M & 11.4 & 3.89 & 26.3 & 5.67 \\
    \rowcolor{gray!12}
    \emph{Ours-3600tokens}                    & 313M & \underline{15.2} & 4.93 & 33.4 & 3.00 \\
    \rowcolor{gray!12}
    \emph{Ours-6000tokens}                    & 313M & \textbf{15.7} & \textbf{5.78} & 36.0 & \textbf{1.67} \\
    \bottomrule
    \end{tabular}
    }
\vspace{-12pt}

\end{table*}

\begin{figure}[h]
    \centering
    \vspace{-10pt}
    \includegraphics[width=0.9\linewidth]{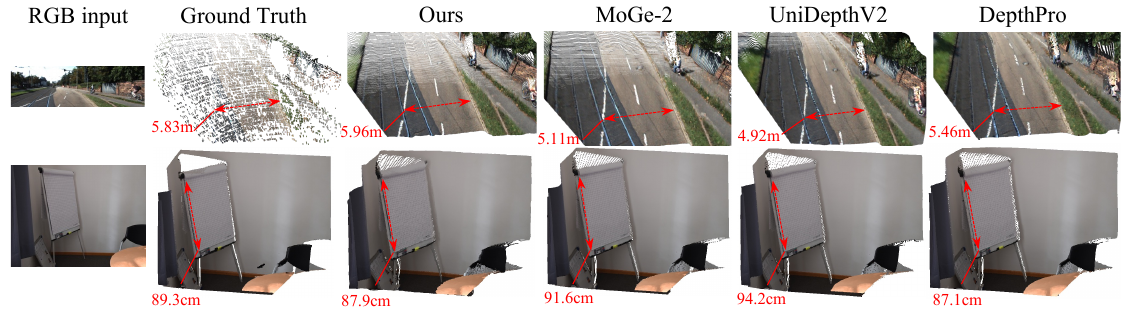}
    \caption{Qualitative metric point-map results on outdoor driving and indoor scenes, spanning depth magnitudes from meters to centimeters. Our model delivers consistent metric accuracy while preserving fine-grained geometric structure and sharp details.}
    \label{fig:vis_compare}
\end{figure}

\vspace{0.04in}\noindent\textbf{Qualitative comparison.}
We compare with leading methods, including MoGe-2~\cite{wang2025moge2}, UniDepthV2~\cite{piccinelli2025unidepthv2}, and DepthPro~\cite{bochkovskiydepth}.
Fig.~\ref{fig:vis_compare} shows that FoundationGeo yields accurate, geometrically consistent metric results across both outdoor driving and indoor scenes, covering depth magnitudes from meters to centimeters. This qualitative stability supports that our spatial fields design and targeted focal-length augmentation enable reliable metric transfer under domain and depth-scale shifts.
\vspace{-10pt}

\subsection{Ablation Study}

We conduct ablations to validate our key design choices and training recipe, including the necessity of two-stage training, spatial fields for metric calibration, field-specific supervision, focal-diverse rendered data, and the impact of ray-direction correction. For the controlled component study in Table~\ref{table:subset_ablation}, all variants use a ViT-Large encoder and the same $\sim$5\% stratified subsets of the training datasets. The single-stage baseline directly optimizes metric geometry for 40K iterations, whereas the two-stage variants first train the relative Base Model for 20K iterations and then optimize metric calibration for another 20K iterations. All controlled experiments are conducted using $8\times$ NVIDIA H20 GPUs. 

We further evaluate the full-data setting in Table~\ref{table:full_ablation} by comparing training from scratch with our two-stage recipe, transferring our rendered data to MoGe-2~\cite{wang2025moge2}, and disentangling the effects of fixed- and diverse-focal rendered data. Finally, we isolate the effect of ray-direction correction with per-dataset experiments (Table~\ref{table:ray_ablation}). Overall, the results show that two-stage training consistently outperforms direct single-stage optimization, spatial fields provide more effective calibration than a global scale, field-specific losses further stabilize metric learning, diverse-focal rendered data improves robustness under camera-intrinsic shifts, and ray-direction correction consistently improves angular accuracy while also yielding better average metric-depth performance.

\begin{table*}[t]
  \centering
  \setlength{\tabcolsep}{3.0pt}
  \renewcommand{\arraystretch}{1.22}

    \begin{minipage}[t]{0.48\textwidth}
      \centering
      \caption{Component ablations averaged over 7 datasets, with ViT-L encoder and 5\% stratified training subsets.}

      \label{table:subset_ablation}
      \resizebox{\linewidth}{!}{
        \begin{tabular}{l|cc}
          \toprule
          \multirow{2}{*}{\textbf{Ablation Variant}}
          & \multicolumn{2}{c}{\textbf{Average}} \\
          & {\small\textit{AbsRel}}$\downarrow$
          & {\small$\delta_1$}$\uparrow$ \\
          \midrule

        Single-stage direct prediction
        & 22.4 & 57.7 \\
        
        Two-stage direct prediction
        & 19.6 & 65.9 \\
        
        Two-stage + global scale
        & 20.1 & 66.8 \\
        
        Two-stage + spatial fields
        & 19.3 & 68.0 \\
        
        \rowcolor{gray!12}
        Two-stage + spatial fields + field losses
        & \textbf{18.8} & \textbf{69.7} \\

          \bottomrule
        \end{tabular}
      }
    \end{minipage}\hfill
    \begin{minipage}[t]{0.48\textwidth}
      \centering
      \caption{Full-data ablations averaged over 7 datasets, FGD denotes our rendered FoundationGeo dataset.}

      \label{table:full_ablation}
    
      {\renewcommand{\arraystretch}{1.18}
      \resizebox{\linewidth}{!}{
        \setlength{\tabcolsep}{2.6pt}
        \begin{tabular}{l|cc|cc|cc}
          \toprule
          \multirow{2}{*}{\textbf{Ablation Variant}}
          & \multicolumn{2}{c|}{\textbf{Average}}
          & \multicolumn{2}{c|}{\textbf{DDAD}}
          & \multicolumn{2}{c}{\textbf{DIODE}} \\
          & {\small\textit{AbsRel}}$\downarrow$
          & {\small$\delta_1$}$\uparrow$
          & {\small\textit{AbsRel}}$\downarrow$
          & {\small$\delta_1$}$\uparrow$
          & {\small\textit{AbsRel}}$\downarrow$
          & {\small$\delta_1$}$\uparrow$ \\
          \midrule

          {Train from Scratch}
          & 15.2 & 78.2
          & -- & --
          & -- & -- \\
    
          \rowcolor{gray!12}
          {Two-stage Training (Ours)}
          & \textbf{14.8} & \textbf{80.8}
          & -- & --
          & -- & -- \\
    
          \midrule
    
          {MoGe-2}
          & 15.7 & 76.8
          & 15.8 & 73.0
          & 17.5 & 66.4 \\
    
          {MoGe-2 + $\mathrm{FGD}$}
          & 15.1 & 79.0
          & 15.8 & 73.8
          & 16.2 & 72.1 \\
    
          \midrule
    
          {w/o FGD}
          & 15.0 & 79.5
          & 20.0 & 69.6
          & 20.5 & 55.2 \\

          {+ Fixed-Focal FGD} 
          & 15.3 & 78.9 
          & 19.2 & 71.7 
          & 18.5 & 63.9 \\
    
          \rowcolor{gray!12}
          \textit{+ $\mathrm{FGD}$ (Ours)}
          & \textbf{14.8} & \textbf{80.8}
          & \textbf{17.6} & \textbf{74.7}
          & \textbf{17.5} & \textbf{69.7} \\
    
          \bottomrule
        \end{tabular}
      }}
    \end{minipage}
\end{table*}

\begin{table}[!t]
  \caption{
  Per-dataset ablation results on metric depth and ray-direction
  estimation.
  For each dataset, we report $\delta_1\uparrow$,
  MaeDeg$\downarrow$, and Pct$_{3^\circ}\uparrow$.
  MaeDeg measures the mean angular error (in degrees) between
  predicted and ground-truth rays, and Pct$_{3^\circ}$ denotes
  the percentage of pixels whose ray-direction error is within
  $3^\circ$.
  $\delta_1$ and Pct$_{3^\circ}$ are reported in percentage.
  }
  \vspace{-4pt}
  \label{table:ray_ablation}
  \centering
  \renewcommand{\arraystretch}{1.25}

  \resizebox{0.97\linewidth}{!}{
    \begin{tabular}{l|ccc|ccc|ccc|ccc|ccc}
      \toprule
      \multirow{2}{*}{\textbf{Ablation}}
      & \multicolumn{3}{c|}{\textbf{NYUv2}}
      & \multicolumn{3}{c|}{\textbf{DIODE}}
      & \multicolumn{3}{c|}{\textbf{DDAD}}
      & \multicolumn{3}{c|}{\textbf{HAMMER}}
      & \multicolumn{3}{c}{\textbf{Avg.}} \\

      & $\delta_1\uparrow$
      & MaeDeg$\downarrow$
      & Pct$_{3^\circ}\uparrow$

      & $\delta_1\uparrow$
      & MaeDeg$\downarrow$
      & Pct$_{3^\circ}\uparrow$

      & $\delta_1\uparrow$
      & MaeDeg$\downarrow$
      & Pct$_{3^\circ}\uparrow$

      & $\delta_1\uparrow$
      & MaeDeg$\downarrow$
      & Pct$_{3^\circ}\uparrow$

      & $\delta_1\uparrow$
      & MaeDeg$\downarrow$
      & Pct$_{3^\circ}\uparrow$ \\
      \midrule

    \textit{w/o Ray Direction Corr.}
    & 92.7 & 1.715 & 86.8
    & 52.6 & 4.232 & 32.8
    & 72.8 & 2.799 & 61.1
    & \textbf{72.9} & 2.810 & 61.3
    & 72.8 & 2.889 & 60.5 \\
    
    \rowcolor{gray!12}
    \textit{\textbf{Ours}}
    & \textbf{93.0} & \textbf{1.484} & \textbf{90.2}
    & \textbf{69.7} & \textbf{2.658} & \textbf{64.0}
    & \textbf{74.7} & \textbf{2.769} & \textbf{61.5}
    & {69.6} & \textbf{2.510} & \textbf{68.2}
    & \textbf{76.8} & \textbf{2.355} & \textbf{71.0} \\

      \bottomrule
    \end{tabular}
  }

  \vspace{-4pt}
\end{table}

\vspace{0.04in}\noindent \textbf{Effectiveness of Spatial Fields (Table~\ref{table:subset_ablation}, row~2 vs.\ row~4):}
Starting from the same two-stage direct-prediction baseline, introducing spatial fields improves the average \textit{AbsRel} from 19.6 to 19.3 and $\delta_1$ from 65.9 to 68.0. This indicates that directly fine-tuning the point-map head with metric supervision remains limited. By explicitly modeling pixel-wise scale and directional corrections, the spatial fields provide dedicated calibration pathways that more effectively bridge affine-invariant geometry and metric prediction.

\vspace{0.04in}\noindent \textbf{Spatial Fields vs.\ Global Scale (Table~\ref{table:subset_ablation}, row~3 vs.\ row~4):}
Under the same two-stage setting, replacing the MoGe-2-style image-level global scale with spatial fields improves the average \textit{AbsRel} from 20.1 to 19.3 and $\delta_1$ from 66.8 to 68.0. Although a single global factor can correct the dominant image-level scale ambiguity, it cannot account for spatially varying scale drift or ray-direction inconsistency across pixels. In contrast, the proposed spatial fields provide locally adaptive scale and directional calibration, enabling more accurate relative-to-metric transfer than global scaling.

\vspace{0.04in} \noindent \textbf{Effectiveness of Fields Loss Design (Table~\ref{table:subset_ablation}, row~4 vs.\ row~5):}
Adding the field-specific objectives improves the average \textit{AbsRel} from 19.3 to 18.8 and $\delta_1$ from 68.0 to 69.7. This suggests that the coupled metric loss alone leaves the decomposition between geometry prediction and field-based calibration under-constrained, since the point-map and field heads may absorb overlapping residual errors. Direct supervision of the scale and ray-direction fields better anchors their respective roles, encouraging them to act as lightweight calibration modules rather than redundant geometry predictors and thereby improving the stability and accuracy of metric calibration.

\vspace{0.04in} \noindent \textbf{Two-stage Training Strategy 
(Table~\ref{table:subset_ablation}, rows~1 vs.\ row~2;
Table~\ref{table:full_ablation}, row~1 vs.\ row~2):}
Two-stage training consistently outperforms direct metric learning at both controlled and full-data scales. Under full-data training, initializing from the Stage-1 relative Base Model further improves the results from 15.2/78.2 to 14.8/80.8 in \textit{AbsRel}/$\delta_1$, compared with training the metric model from scratch. These results indicate that separating affine-invariant geometry learning from metric calibration provides a stronger initialization: the first stage establishes a stable structural prior, allowing the second stage to focus on lightweight metric calibration while preserving geometric fidelity.

\vspace{0.04in} \noindent \textbf{Effectiveness of Render Data (Table~\ref{table:full_ablation}, rows~3--7):}
Rendered data provides a targeted adaptation signal under camera and domain shifts. Fine-tuning an off-the-shelf metric model with our renders improves robustness, indicating that the synthetic set effectively reduces intrinsic mismatch rather than merely adding more samples. To separate focal-length diversity from synthetic data scale-up, we further construct a controlled Fixed-Focal FGD with the focal length fixed at 1320\,px using the same scenes and camera trajectories. Although the fixed-focal data improves performance on DDAD and DIODE, it slightly degrades the overall average. In contrast, Diverse-Focal FGD delivers more consistent gains across datasets and further improves both camera-sensitive benchmarks. These results show that merely adding rendered data from a single camera regime is insufficient for broad generalization, whereas diverse focal-length coverage more effectively improves robustness under camera-intrinsic shifts.

\vspace{0.04in} \noindent \textbf{Effectiveness of Ray Direction Correction (Table~\ref{table:ray_ablation}, row~1 vs. row~2):}
Ray-direction correction consistently improves the ray-centric metrics across all evaluated datasets, indicating that directional bias is a distinct source of point-map error that cannot be resolved by scale calibration alone. The gains are particularly evident under stronger camera and domain shifts, where inaccurate ray directions can produce substantial 3D inconsistency even when depth scale is reasonably calibrated. The clear improvement in average $\delta_1$ further suggests that more accurate ray geometry generally benefits metric-depth prediction. Although the per-dataset $\delta_1$ is not uniformly improved, the ray metrics remain consistently better; this discrepancy may partly arise from optimization interference between the shared backbone and multiple prediction heads.

\section{Conclusion}
\label{sec:conclusion}

We present FoundationGeo, a two-stage framework that bridges affine-invariant relative geometry and monocular metric prediction. In the first stage, a DINOv3-initialized backbone is trained on a curated 10.2M-image corpus to learn globally consistent and detail-preserving relative geometry. In the second stage, lightweight pixel-wise calibration fields correct ray-direction bias and spatially varying scale drift, converting the relative point map into metrically consistent 3D geometry. We further identify camera-intrinsic coverage, particularly focal-length distribution mismatch, as a major bottleneck for zero-shot metric generalization, and address it through targeted Blender-based augmentation with 23,700 images spanning under-covered focal regimes.
Extensive zero-shot evaluations demonstrate that FoundationGeo achieves the best overall metric-depth performance across seven benchmarks without requiring ground-truth camera intrinsics at inference. Meanwhile, the Stage-1 model provides a strong affine-invariant geometry prior, achieving the best overall relative-depth performance and competitive boundary sharpness. These results suggest that robust monocular metric geometry depends not only on stronger relative representations, but also on locally adaptive calibration and principled camera-distribution coverage. Future work may extend this framework to temporal or multi-view observations, where geometric consistency across frames can further reduce monocular ambiguity and improve robustness under unseen camera models.

\section{Acknowledgment}
This work was conducted during an internship at Voyager Research, DiDi Chuxing. The research was supported by the Hong Kong Research Grants Council (RGC) through the General Research Fund (Grants No. 17202422, 17212923, and 17215025), the Theme-based Research Scheme (Grant No. T45-701/22-R), and the Strategic Topics Grant (Grant No. STG3/E-605/25-N). Additionally, part of this research was conducted at the JC STEM Lab of Robotics for Soft Materials, funded by The Hong Kong Jockey Club Charities Trust.

\bibliographystyle{splncs04}
\bibliography{main}

\clearpage
\appendix
\setcounter{page}{1}
\section*{Supplementary Material}
\addcontentsline{toc}{section}{Supplementary Material}

This supplementary material provides additional implementation details, algorithmic explanations, and dataset descriptions to complement the main paper. Sec.~\ref{sec:loss} details the training objectives used in both stages, including the affine-invariant relative-geometry losses and the Stage-II hyperparameter settings for metric calibration. Sec.~\ref{sec:ray_details} presents the detailed algorithm for ray-direction correction, clarifying how the local tangent basis is constructed and how bounded angular offsets are applied in practice. Sec.~\ref{sec:FGD} introduces the FoundationGeo Dataset and the underlying Blender-based data engine, including scene composition, camera configuration, and manually designed trajectory layouts. Sec.~\ref{sec:experiments_detail} provides additional experimental details, including the unified evaluation protocol and representative examples of training-data filtering. Finally, Sec.~\ref{sec:future} discusses current limitations and outlines several promising directions for future work.

\section{Detailed Loss Design}
\label{sec:loss}

\subsection{Stage-I: Relative Geometry Objective}

We adopt a set of affine-invariant and geometry-aware losses to train a strong relative geometry base model. In Stage-I, the network predicts an affine-invariant point map $\hat{\mathbf P}\in\mathbb{R}^{H\times W\times 3}$ and a reliability mask $\hat{\mathbf M}\in[0,1]^{H\times W}$, and is optimized with complementary global, local, normal, edge, and mask supervision.

\vspace{0.05in}\noindent \textbf{Global Alignment Loss.}
The global alignment loss fits $\hat{\mathbf P}$ to ground-truth 3D points under a global scale--shift ambiguity. Let $\hat{\mathbf p}_i$ denote the predicted 3D point for the $i$-th pixel and $\mathbf p_i$ its corresponding ground truth. The global affine-invariant loss is defined as
\begin{equation}
\mathcal L_{\rm global}
=
\sum_{i\in \mathcal M}
\frac{1}{z_i}
\left\|
s^*\hat{\mathbf p}_i+\mathbf t^*-\mathbf p_i
\right\|_1,
\label{eq:global_loss_sup}
\end{equation}
where $(s^*,\mathbf t^*)$ are the alignment parameters that transform the predicted affine-invariant point map into the ground-truth camera space, and $\mathcal M$ is the valid supervision mask. In practice, $(s^*,\mathbf t^*)$ are estimated by solving a global alignment between prediction and ground truth. The weighting term ${1}/{z_i}$, where $z_i$ is the $z$-coordinate of $\mathbf p_i$, balances the supervision strength across large depth ranges.

This objective encourages the model to recover globally consistent affine-invariant geometry while remaining agnostic to absolute metric scale. When combined with the local losses below, it provides a coarse-to-fine training signal for relative geometry learning.

\vspace{0.05in}\noindent \textbf{Multi-scale Local Patch Losses.}
To preserve local structures and high-frequency details, we additionally apply multi-scale local patch supervision. The overall local objective is
\begin{equation}
\mathcal L_{\rm local}
=
\sum_{\alpha \in \mathcal A}
\mathcal L_{S(\alpha)},
\label{eq:local_loss_total_sup}
\end{equation}
where $\mathcal A$ is the set of neighborhood scales.

For each scale $\alpha$, we first construct a local spherical neighborhood around an anchor point $\mathbf p_j$:
\begin{equation}
\mathcal S_j
=
\left\{
i \;\middle|\;
\|\mathbf p_i-\mathbf p_j\|\le r_j,\; i\in\mathcal M
\right\},
\label{eq:sphere_region_sup}
\end{equation}
with radius
\begin{equation}
r_j
=
\alpha \cdot z_j \cdot \frac{\sqrt{W^2+H^2}}{2f},
\label{eq:radius_sup}
\end{equation}
where $z_j$ is the depth of $\mathbf p_j$, $f$ is the ground-truth focal length, and $(W,H)$ is the image resolution. This design makes the neighborhood size adapt to both scene depth and camera intrinsics.

A local affine alignment is then solved within each spherical patch using scale and translation parameters $(s_j^*,\mathbf t_j^*)$, and the patch loss is defined as
\begin{equation}
\mathcal L_{S(\alpha)}
=
\sum_{j\in\mathcal H_\alpha}
\sum_{i\in\mathcal S_j}
\frac{1}{z_i}
\left\|
s_j^* \hat{\mathbf p}_i + \mathbf t_j^* - \mathbf p_i
\right\|_1,
\label{eq:local_loss_sup}
\end{equation}
where $\mathcal H_\alpha$ denotes the set of anchors sampled at scale $\alpha$.

In practice, we use different local scales for different supervision sources. Synthetic data uses three levels $\mathcal A_{\rm syn}=\{\tfrac{1}{4}, \tfrac{1}{16}, \tfrac{1}{64}\}$, SfM data uses $\mathcal A_{\rm sfm}=\{\tfrac{1}{4}, \tfrac{1}{16}\}$, and LiDAR data uses only $\mathcal A_{\rm lidar}=\{\tfrac{1}{4}\}$.

\noindent \textbf{Surface-Normal Consistency Loss.}
To encourage piecewise smooth yet detail-preserving geometry, we impose a surface-normal consistency loss:
\begin{equation}
\mathcal L_{\rm normal}
=
\sum_{i\in\mathcal M}
\angle(\hat{\mathbf n}_i,\mathbf n_i),
\label{eq:normal_loss_sup}
\end{equation}
where $\hat{\mathbf n}_i$ and $\mathbf n_i$ denote the predicted and ground-truth surface normals, respectively, and $\angle(\cdot,\cdot)$ measures their angular discrepancy.

\noindent \textbf{Edge Loss.}
To better preserve geometric discontinuities and local structure, we further supervise the directions of neighboring 3D point differences. Let
\begin{equation}
\Delta_x \hat{\mathbf p}_{u,v} = \hat{\mathbf p}_{u,v} - \hat{\mathbf p}_{u+1,v},
\qquad
\Delta_y \hat{\mathbf p}_{u,v} = \hat{\mathbf p}_{u,v} - \hat{\mathbf p}_{u,v+1},
\end{equation}
denote the horizontal and vertical edge vectors of the predicted point map, and define $\Delta_x \mathbf p_{u,v}$ and $\Delta_y \mathbf p_{u,v}$ similarly for the ground truth. We then penalize their angular discrepancy:
\begin{equation}
\mathcal L_{\rm edge}
=
\sum_{(u,v)\in\mathcal M_x}
\angle(\Delta_x \hat{\mathbf p}_{u,v}, \Delta_x \mathbf p_{u,v})
+
\sum_{(u,v)\in\mathcal M_y}
\angle(\Delta_y \hat{\mathbf p}_{u,v}, \Delta_y \mathbf p_{u,v}),
\label{eq:edge_loss_sup}
\end{equation}
where $\mathcal M_x$ and $\mathcal M_y$ denote the valid neighboring pixel pairs in the horizontal and vertical directions, respectively. This loss encourages the predicted point map to preserve local edge directions and sharp geometric transitions.

\vspace{0.05in}\noindent \textbf{Mask Supervision Loss.}
To suppress unreliable regions, we supervise the predicted reliability mask with
\begin{equation}
\mathcal L_{\rm mask}
=
\left\|
\hat{\mathbf M} - \left(1-\mathbf M_{\rm inf}\right)
\right\|_2^2,
\label{eq:mask_loss_sup}
\end{equation}
where $\mathbf M_{\rm inf}$ denotes the invalid-region indicator derived from ground-truth labeling.

\noindent \textbf{Overall Stage-I Objective.}
The final Stage-I loss is label-type dependent:
\begin{equation}
\mathcal L_{\rm relative}^{(t)}
=
\lambda_{\rm g}^{(t)}\mathcal L_{\rm global}
+
\sum_{\alpha\in\mathcal A_t}
\lambda_{\alpha}^{(t)}\mathcal L_{S(\alpha)}
+
\lambda_{\rm n}^{(t)}\mathcal L_{\rm normal}
+
\lambda_{\rm e}^{(t)}\mathcal L_{\rm edge}
+
\lambda_{\rm m}^{(t)}\mathcal L_{\rm mask},
\label{eq:loss_stage1_sup}
\end{equation}
where $t\in\{\texttt{synthetic},\texttt{sfm},\texttt{lidar}\}$ denotes the label type.

In all experiments, we set the loss weights as follows:
\begin{equation}
(\lambda_{\rm g},\lambda_{1/4},\lambda_{1/16},\lambda_{1/64},\lambda_{\rm n},\lambda_{\rm e},\lambda_{\rm m})
=
\begin{cases}
(1,\,1,\,1,\,1,\,0.1,\,1,\,0.1), & t=\texttt{synthetic},\\
(1,\,1,\,1,\,0,\,0.1,\,1,\,0.1), & t=\texttt{sfm},\\
(1,\,1,\,0,\,0,\,0,\,0,\,0), & t=\texttt{lidar}.
\end{cases}
\label{eq:stage1_weights}
\end{equation}
Unless otherwise stated, the global and local alignment terms all use unit weight. In implementation, the global loss uses an alignment resolution of $48$, while the local patch losses at levels $\{4,16,64\}$ use alignment resolutions $\{24,12,6\}$ and numbers of sampled patches $\{16,256,4096\}$, respectively.

\subsection{Stage-II Training Details}

As described in the main paper, Stage-II retains the relative-geometry supervision from Stage-I and further introduces coupled metric regression together with decoupled supervision on the ray-direction correction field and the pixel-wise scale field. Here we only provide the implementation details and hyperparameter settings used in training.

\vspace{0.05in}\noindent \textbf{Overall Stage-II Objective.}
Following the formulation in the main paper, the overall Stage-II training objective is
\begin{equation}
\mathcal L_{\rm FoundationGeo}
=
\mathcal L_{\rm relative}
+
\mathcal L_{\rm metric}
+
\gamma_{\rm s}\mathcal L_{\rm scalefield}
+
\gamma_{\rm r}\mathcal L_{\rm ray}
+
\gamma_{\Delta}\mathcal L_{\Delta},
\label{eq:overall_obj_sup}
\end{equation}
where $\mathcal L_{\rm relative}$ denotes the source-dependent structural supervision inherited from Stage-I, including the enabled global, local, normal, and mask terms for each supervision source. In our implementation, we set $\gamma_{\rm s}=0.2$, $\gamma_{\rm r}=0.1$, and $\gamma_{\Delta}=0.05$.

For synthetic data, we use the full Stage-I structural regularization, including the global loss, local patch losses at levels $\{4,16,64\}$, normal loss, and mask loss, and combine them with the Stage-II metric calibration terms. For SfM data, we use global loss, local patch losses at levels $\{4,16\}$, normal loss, and mask loss. For LiDAR data, we use only the global loss, the level-$4$ local patch loss, and the metric calibration terms.

The metric loss uses unit weight, while the additional calibration terms are weighted by $\gamma_{\rm s}=0.2$, $\gamma_{\rm r}=0.1$, and $\gamma_{\Delta}=0.05$. The normal loss and mask loss are weighted by $1.0$ and $0.1$, respectively, when enabled through $\mathcal L_{\rm relative}$.

For the scale-field supervision, we apply the loss in the log domain and clamp the target scale into the range $[0.05,\,20.0]$. The robust penalty uses $\beta=\log(1.25)$, corresponding to an approximately $\pm25\%$ tolerance in linear scale. For the ray-direction loss, we use a robust angular penalty with $\beta=3^\circ$, and clamp the angular error to the range $[0.05^\circ,\,30^\circ]$. For the correction-field regularization, we use $q=2$.

\begin{table}[h]
\centering
\small
\caption{Stage-II loss configuration for different supervision sources.}
\label{tab:stage2_loss_config}
\begin{tabular}{lccccccc}
\toprule
Type & Global & Patch-4 & Patch-16 & Patch-64 & Normal & Metric terms & Mask \\
\midrule
Synthetic & \checkmark & \checkmark & \checkmark & \checkmark & \checkmark & \checkmark & \checkmark \\
SfM       & \checkmark & \checkmark & \checkmark &            & \checkmark & \checkmark & \checkmark \\
LiDAR     & \checkmark & \checkmark &            &            &            & \checkmark & \checkmark \\
\bottomrule
\end{tabular}
\end{table}

Overall, this design preserves the strong affine-invariant supervision of Stage-I while explicitly disentangling metric calibration into a pixel-wise scale field and a ray-direction refinement field.


\section{Ray-Direction Correction Details}
\label{sec:ray_details}

\begin{figure}[!h] 
    \centering 
    \includegraphics[width=0.99\linewidth]{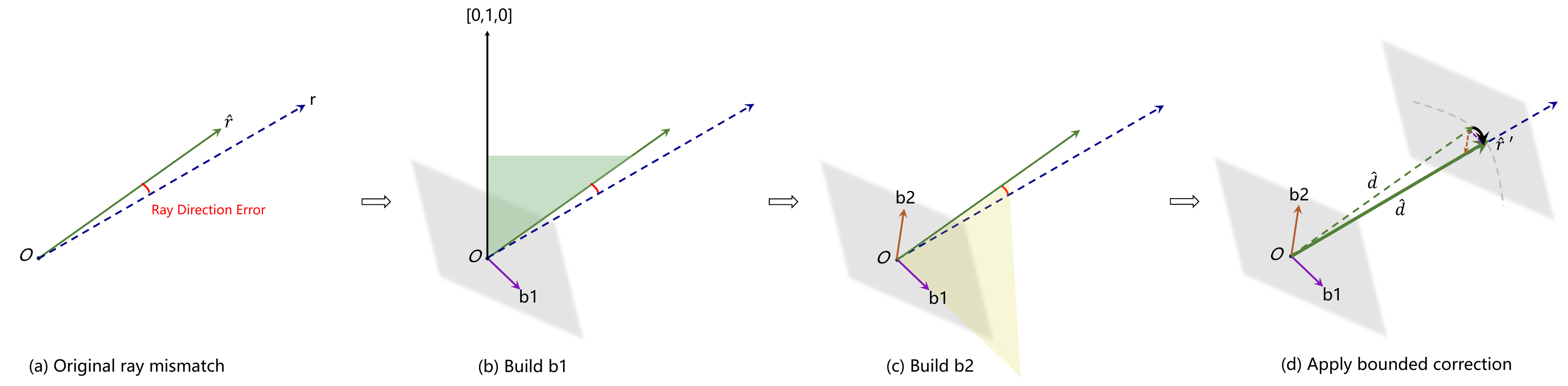} 
    \caption{Illustration of the proposed ray-direction correction. 
    (a) The predicted ray direction may deviate from the target direction, producing a directional error. 
    (b) A stable reference axis is selected to construct the first tangent direction $\mathbf b_1$. 
    (c) The second tangent direction $\mathbf b_2$ is then obtained to form a local orthonormal basis on the tangent plane. 
    (d) Bounded 2D offsets are applied in the tangent plane to correct the ray direction, while preserving the original range $\hat d$.}
    \label{fig:ray_corre} 
\end{figure}

We provide the detailed procedure of the proposed ray-direction correction in Algorithm~\ref{alg:ray_direction_correction}, and illustrate its geometric intuition in Fig.~\ref{fig:ray_corre}. Given a predicted affine-invariant point map $\hat{\mathbf P}$, we decompose each point $\hat{\mathbf p}_i$ into its range $\hat d_i=\|\hat{\mathbf p}_i\|_2$ and unit ray direction $\hat{\mathbf r}_i=\hat{\mathbf p}_i/\hat d_i$. As shown in Fig.~\ref{fig:ray_corre}, we then construct a stable local tangent basis $(\mathbf b_{1,i}, \mathbf b_{2,i})$ orthogonal to $\hat{\mathbf r}_i$ by choosing a reference axis that is not nearly parallel to the current ray direction. The predicted 2D correction is mapped onto this tangent plane, bounded by a $\tanh$ parameterization, and applied to update the ray direction. Finally, the corrected ray is re-normalized and multiplied by the original range to obtain the direction-corrected point $\hat{\mathbf p}_i'$. This design makes the correction explicitly angular and scale-preserving, which helps disentangle directional bias from pixel-wise metric scaling.

\begin{algorithm}[!h]
\caption{Ray-direction correction}
\label{alg:ray_direction_correction}
\begin{algorithmic}
    \State \textbf{input:} relative point map $\hat{\mathbf P}$, correction field $\hat{\Delta}=(\hat\Delta_1,\hat\Delta_2)$
    \State \textbf{output:} corrected point map $\hat{\mathbf P}'$
\Function{ApplyDeltaToRay}{$\hat{\mathbf P}, \hat{\Delta}, \delta_{\max}$}
    \For{each pixel $i$}
        \State $\hat d_i \gets \|\hat{\mathbf p}_i\|_2,\qquad \hat{\mathbf r}_i \gets \hat{\mathbf p}_i / \hat d_i$
        \State set candidate axes $\mathbf a_1=(0,1,0)$ and $\mathbf a_2=(0,0,1)$
        \If{$|\hat{\mathbf r}_i^\top \mathbf a_1| > 0.95$}
            \State $\mathbf a_i \gets \mathbf a_2$
        \Else
            \State $\mathbf a_i \gets \mathbf a_1$
        \EndIf
        \State $\mathbf b_{1,i} \gets \mathrm{norm}(\mathbf a_i \times \hat{\mathbf r}_i)$
        \State $\mathbf b_{2,i} \gets \mathrm{norm}(\hat{\mathbf r}_i \times \mathbf b_{1,i})$
        \State $d_{1,i} \gets \delta_{\max}\tanh(\hat\Delta_{1,i}),\qquad d_{2,i} \gets \delta_{\max}\tanh(\hat\Delta_{2,i})$
        \State $\tilde{\mathbf r}_i \gets \hat{\mathbf r}_i + d_{1,i}\mathbf b_{1,i} + d_{2,i}\mathbf b_{2,i}$
        \State $\hat{\mathbf r}'_i \gets \mathrm{norm}(\tilde{\mathbf r}_i)$
        \State $\hat{\mathbf p}'_i \gets \hat d_i \hat{\mathbf r}'_i$
    \EndFor
    \State \Return $\hat{\mathbf P}'$
\EndFunction
\end{algorithmic}
\end{algorithm}

Algorithm~\ref{alg:ray_direction_correction} and Fig.~\ref{fig:ray_corre} summarize the ray-direction correction procedure used in Stage-II. The key idea is to modify only the ray direction while keeping the predicted range unchanged, so that the correction focuses on directional bias rather than absorbing metric scale. The tangent-space parameterization also makes the update geometrically interpretable, as the two predicted offsets correspond to small angular perturbations along two orthogonal directions on the local tangent plane. This design has three desirable properties: it is \emph{scale-preserving}, since the original range is left unchanged; it is \emph{geometrically interpretable}, since the correction is explicitly represented as bounded 2D offsets in the tangent plane; and it is \emph{numerically stable}, since the dynamic reference-axis selection avoids degenerate cross products when the ray is close to a fixed canonical axis. In addition, the bounded $\tanh$ parameterization prevents overly large corrections during training.

\section{FoundationGeo Dataset}
\label{sec:FGD}

To improve camera-model diversity in metric training, we build a targeted synthetic dataset in Blender, referred to as the \textbf{FoundationGeo Dataset}. Its main purpose is to complement the intrinsic distribution of the original training corpus with controllable rendered data, especially in under-covered focal regimes. Unlike passive aggregation from existing datasets, our data engine allows joint control of scene layout, camera trajectories, and focal settings while preserving accurate geometric supervision. In total, the rendered pool contains 23,700 images, including 22,900 main training images across seven scenes and a 400-image object-centric subset from one indoor scene; an additional 400 rendered images are not used in training.

\begin{figure}[!ht] 
    \centering 
    \includegraphics[width=0.99\linewidth]{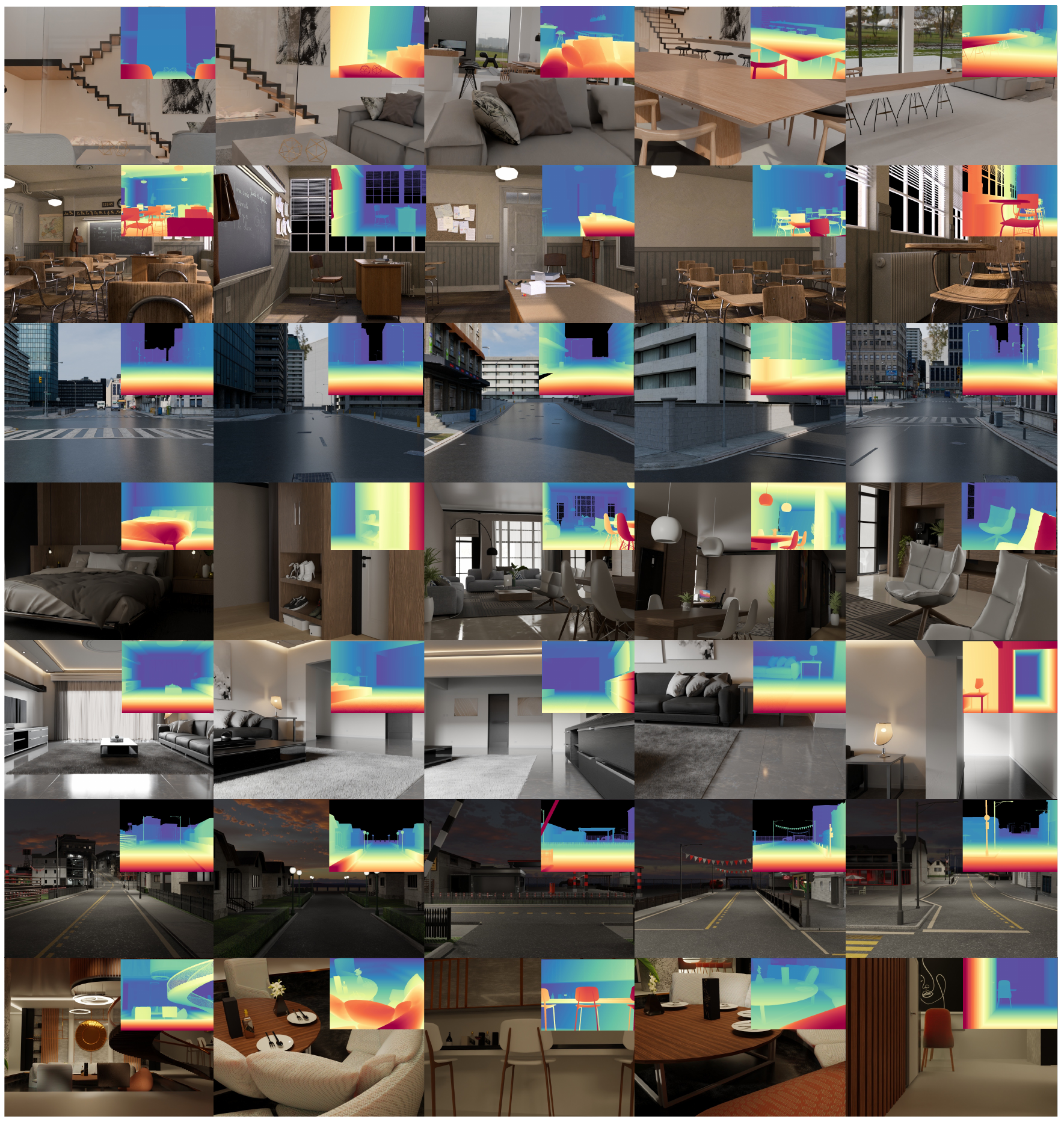} 
    \caption{Overview of the FoundationGeo Dataset. We build a Blender-based synthetic data engine with seven scenes, including five indoor scenes and two outdoor scenes. The figure shows representative RGB images and corresponding depth maps from each scene, illustrating the diversity of layouts, viewpoints, and geometric structures covered by the dataset.}
    \label{fig:FGD} 
\end{figure}

\noindent \textbf{Tools and Assets.}
The FoundationGeo Dataset is built with a Blender-based data engine and contains seven scenes in total, including five indoor scenes and two outdoor scenes, spanning room-scale, building-scale, and open-environment layouts. For each frame, we export aligned RGB images, depth maps, focal metadata, and camera poses, all indexed consistently for direct correspondence and easy integration into metric training. Camera poses are stored as absolute camera-to-world transformations. Figure~\ref{fig:FGD} shows representative RGB and depth examples from the seven scenes, illustrating the diversity of scene structure, scale, and geometry covered by the rendered data engine.

\noindent \textbf{Camera Configuration.}
All images are rendered at a resolution of $1024\times768$ with a standard $4{:}3$ aspect ratio. Rather than fixing a single intrinsic setup, we render the dataset under varying focal lengths to increase camera diversity and better cover under-represented intrinsic regimes. This makes the dataset a targeted metric-stage supplement rather than generic synthetic augmentation.

\noindent \textbf{Layout Configuration.}
The seven scenes are designed to cover both indoor and outdoor geometry distributions. The scene-level image counts are: Scene 1 (indoor, 3,300), Scene 2 (indoor, 1,500), Scene 3 (outdoor, 8,700), Scene 4 (indoor, 3,000), Scene 5 (indoor, 1,000), Scene 6 (outdoor, 3,000), and Scene 7 (indoor, 2,400). In addition, one indoor scene contains a 400-image object-centric subset for controlled rendering analysis.

All camera trajectories are manually annotated. Instead of relying on fully random camera sampling, we explicitly design trajectories and select valid viewpoints to ensure useful geometric coverage, stable perspective variation, and visually meaningful observations. This improves supervision quality by avoiding degenerate views and invalid scene configurations, while also producing perspective changes and structural depth transitions that are beneficial for metric calibration. Overall, although modest in scale compared with the full multi-source corpus, the FoundationGeo Dataset is intentionally designed as a targeted supplement that injects clean geometric supervision under controlled camera settings missing from the original training distribution.

\section{Experimental Details}
\label{sec:experiments_detail}
\subsection{Evaluation Protocol Details}

In monocular depth estimation, widely used benchmarks such as NYUv2 and KITTI are often evaluated under heterogeneous preprocessing protocols, making fair comparison across different works difficult. In particular, the same model can yield noticeably different results depending on how each paper crops, downsamples, or filters the data. To mitigate this issue, we adopt the unified evaluation protocol of MoGe~\cite{wang2025moge} and follow its benchmark datasets: NYUv2~\cite{silberman2012indoor}, KITTI~\cite{Uhrig2017THREEDV}, ETH3D~\cite{schops2019bad}, iBims-1~\cite{koch2018evaluation,koch2020comparison}, Sintel~\cite{butler2012naturalistic}, DDAD~\cite{guizilini20203d}, DIODE~\cite{vasiljevic2019diode}, Spring~\cite{mehl2023spring}, and HAMMER~\cite{jung2023importance}. Among these, Spring is excluded from evaluation because it is used for training, and Sintel is used only for relative-depth evaluation due to the absence of metric scale. As a result, we report relative depth on eight datasets, and metric depth on the seven datasets with reliable metric annotations.

All datasets undergo dataset-specific preprocessing to enforce a consistent and robust evaluation setting. Typical steps include resolution normalization, such as center-cropping KITTI and Sintel to a fixed aspect ratio or downsampling the high-resolution ETH3D images, as well as systematic data cleaning to mitigate sensor noise and ground-truth artifacts. For NYUv2 and DIODE, unreliable boundary regions are removed using edge-based filtering; for NYUv2 in particular, depths beyond 5\,m are discarded and reflective surfaces are manually masked. Synthetic and video datasets require additional handling: sky regions are masked in Sintel, and DDAD images are cropped to remove visible parts of the ego-vehicle. Overall, this standardized preprocessing reduces evaluation ambiguity and ensures fairer comparison across methods; further implementation details can be found in the supplementary material of MoGe~\cite{wang2025moge}.

\subsection{Training Data Filtering}

We collect a large-scale multi-source training set containing 10.2M samples. Since not all collected samples are suitable for training, we apply a filtering process to remove images with severe domain mismatch, weak geometric cues, or unreliable annotations. The overall filtering strategy is described in Sec.~\ref{3.1}; here we provide several representative examples.

\begin{figure}[ht]
    \centering
    \includegraphics[width=0.55\linewidth]{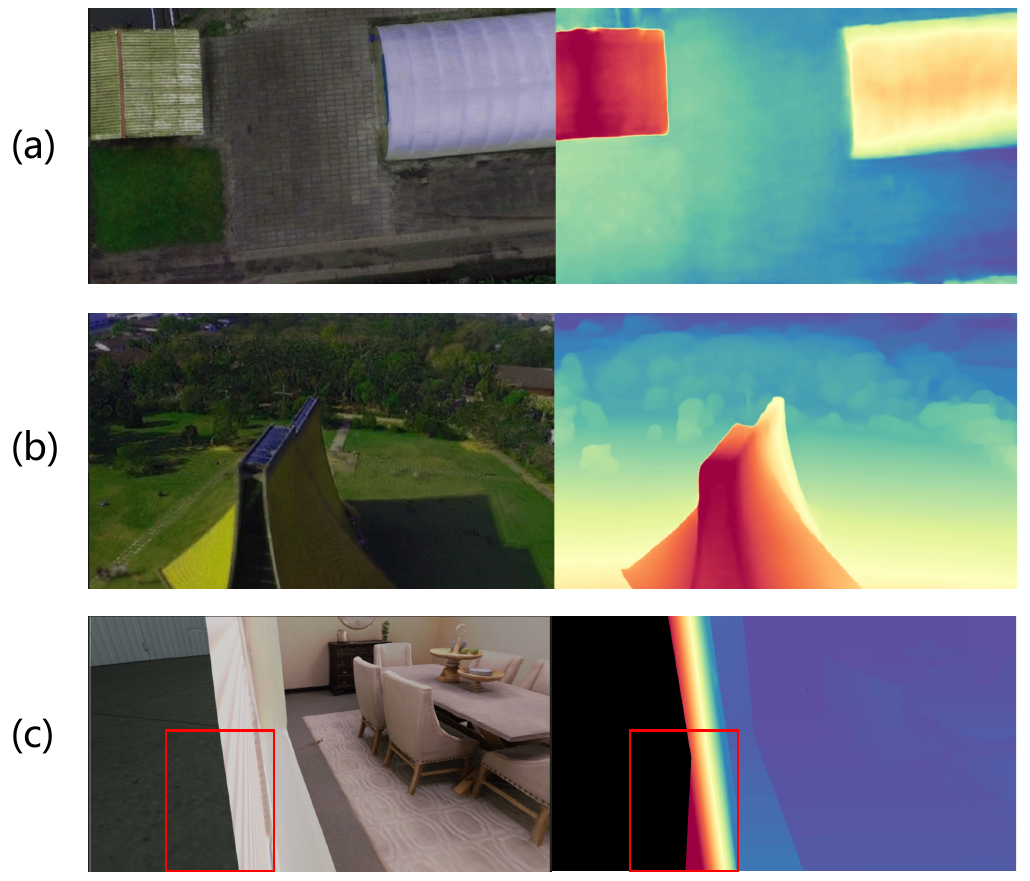}
    \caption{Representative examples of data filtering. 
    (a) Bird’s-eye-view or remote-sensing-style samples are removed due to severe domain mismatch. 
    (b) Samples with clear perspective and near--far depth ordering are retained. 
    (c) Mislabeled samples with scene-boundary violations and inconsistent depth annotations are filtered out.}
    \label{fig:filter}

\end{figure}

As shown in Fig.~\ref{fig:filter}, we discard samples that fall far outside the target domain or contain inconsistent geometry labels, and retain samples with reliable perspective and depth structure. In particular, bird’s-eye-view or remote-sensing-style samples, such as the example in Fig.~\ref{fig:filter}(a), are removed because their imaging geometry differs substantially from the target distribution of our model. In contrast, samples such as Fig.~\ref{fig:filter}(b) are retained because they exhibit clear near--far depth ordering and informative perspective cues. We also remove mislabeled samples such as Fig.~\ref{fig:filter}(c), where the camera extends beyond the scene boundary and the invalid out-of-bound region is not properly masked, leading to a clear mismatch between RGB content and depth annotation.

\section{Limitations and Future Work}
\label{sec:future}

\noindent \textbf{Camera-model coverage beyond focal length.}
Our study identifies focal-length distribution mismatch as a major bottleneck for zero-shot metric generalization, and shows that targeted rendered data can effectively improve robustness under such intrinsic shifts. However, the current intervention mainly focuses on focal-length coverage and does not fully span the broader space of camera models encountered in real applications, such as variations in principal point, aspect ratio, sensor size, distortion, or other imaging characteristics. Extending the current data engine and training strategy to cover a richer set of intrinsic factors is an important direction for future work.

\noindent \textbf{Capacity of lightweight spatial calibration fields.}
Our ray-direction correction field and pixel-wise scale field are intentionally designed as lightweight calibration modules on top of a strong relative backbone. This design is effective and stable, but it also limits the correction capacity under more extreme out-of-distribution settings, such as severe camera shifts, highly unusual scene geometry, or strong appearance degradations. In future work, it would be interesting to explore more expressive yet still well-constrained calibration designs, such as hierarchical spatial fields, uncertainty-aware calibration, or stronger coupling between global camera cues and local geometric refinement.

\noindent \textbf{Scale and diversity of the synthetic data engine.}
Although the FoundationGeo Dataset provides a useful targeted supplement for under-covered camera regimes, it is still relatively limited in scene scale and asset diversity compared with the complexity of real-world open-domain data. In particular, our current rendered set is built from seven scenes with manually designed valid trajectories, which provides clean supervision but does not yet exhaust the diversity of layouts, objects, materials, and motion patterns seen in practice. A promising future direction is to build a larger and more automated data engine with broader scene coverage, richer controllable camera parameters, and more scalable trajectory generation, so that synthetic data can serve as a stronger and more systematic tool for metric-geometry generalization.

\end{document}